\documentclass[times,twocolumn,final,longtitle]{elsarticle}
\usepackage{medima}
\usepackage{framed,multirow}

\usepackage{amssymb}
\usepackage{latexsym}

\usepackage{url}
\usepackage{xcolor}

\usepackage{hyperref}

\usepackage{bm}
\usepackage{graphicx}
\usepackage{amsmath}

\DeclareMathOperator*{\argmin}{arg\,min}
\usepackage{xspace}
\usepackage{placeins}
\usepackage{booktabs}

\newtheorem{definition}{Definition}

\definecolor{newcolor}{rgb}{.8,.349,.1}

\journal{Medical Image Analysis}

\begin{document}

\verso{Dieuwertje Alblas \textit{et~al.}}

\begin{frontmatter}

\title{SIRE: scale-invariant, rotation-equivariant estimation of artery orientations using graph neural networks} % Centerline Extraction using Graph Neural Networks} %\tnoteref{tnote1}}%
%\tnotetext[tnote1]{This is an example for title footnote coding.}

\author[1]{Dieuwertje \snm{Alblas}\corref{cor1}}
\cortext[cor1]{Corresponding author: 
  E-mail: d.alblas@utwente.nl}
\author[1]{Julian \snm{Suk}}
\author[1]{Christoph \snm{Brune}} %\fnref{fn1}}
%\fntext[fn1]{This is author footnote for second author.}
\author[2,3]{Kak Khee \snm{Yeung}}
\author[1]{Jelmer M. \snm{Wolterink}}
% %% Third author's email
% \ead{author3@author.com}

\address[1]{Department of Applied Mathematics, Technical Medical Centre, University of Twente, Drienerlolaan 5 7522 NB Enschede, The Netherlands}
\address[2]{Amsterdam UMC location Vrije Universiteit Amsterdam, Department of Surgery, De Boelelaan 1117 1081 HV Amsterdam, The Netherlands}
\address[3]{Amsterdam Cardiovascular Sciences, Microcirculation, Amsterdam, The Netherlands}

\received{November 7, 2023 \\ \textit{Preprint, under review}}
% \finalform{XXX 2023}
% \accepted{XXX 2023}
% \availableonline{XXX 2023}
\communicated{D. Alblas}

\begin{abstract}
The orientation of a blood vessel as visualized in 3D medical images is an important descriptor of its geometry that can be used for centerline extraction and subsequent segmentation, labelling, and visualization. Blood vessels appear at multiple scales and levels of tortuosity, and determining the exact orientation of a vessel is a challenging problem. Recent works have used 3D convolutional neural networks (CNNs) for this purpose, but CNNs are sensitive to variations in vessel size and orientation. We present SIRE: a scale-invariant rotation-equivariant estimator for local vessel orientation. SIRE is modular and has strongly generalising properties due to symmetry preservations.

SIRE consists of a gauge equivariant mesh CNN (GEM-CNN) that operates in parallel on multiple nested spherical meshes with different sizes. The features on each mesh are a projection of image intensities within the corresponding sphere. These features are intrinsic to the sphere and, in combination with the gauge equivariant properties of GEM-CNN, lead to SO(3) rotation equivariance. Approximate scale invariance is achieved by weight sharing and use of a symmetric maximum aggregation function to combine predictions at multiple scales.  Hence, SIRE can be trained with arbitrarily oriented vessels with varying radii to generalise to vessels with a wide range of calibres and tortuosity.

We demonstrate the efficacy of SIRE using three datasets containing vessels of varying scales; the vascular model repository (VMR), the ASOCA coronary artery set, and an in-house set of abdominal aortic aneurysms (AAAs). We embed SIRE in a centerline tracker which accurately tracks large calibre AAAs, regardless of the data SIRE is trained with. Moreover, a tracker can use SIRE to track small-calibre tortuous coronary arteries, even when trained only with large-calibre, non-tortuous AAAs. Additional experiments are performed to verify the rotational equivariant and scale invariant properties of SIRE. 

In conclusion, by incorporating SO(3) and scale symmetries, SIRE can be used to determine orientations of vessels outside of the training domain, offering a robust and data-efficient solution to geometric analysis of blood vessels in 3D medical images.
\end{abstract}


\end{frontmatter}

%\linenumbers

\section{Introduction}
Cardiovascular diseases are the leading cause worldwide of mortality and morbidity \citep{abubakar2015global}. In Europe alone, an estimated 113 million people suffered from cardiovascular diseases in 2019 \citep{timmis2022european}. Proper diagnosis, prognosis and treatment planning of cardiovascular diseases require personalised 3D vascular models, that can be used for a range of downstream tasks. For example, vessel diameters can be extracted from these vascular models, which can be used for, e.g., monitoring of abdominal aortic aneurysms (AAAs) \citep{vaitenas2023abdominal, wanhainen2019editor}, or quantification of stenosis in coronary arteries \citep{shahzad2013automatic, kiricsli2013standardized}. Moreover, personalised 3D models can be used to calculate hemodynamic parameters, such as wall shear stress and blood flow velocity using computational fluid dynamics (CFD) \citep{taylor2023patient}. Furthermore, vessel curvature can be measured from these models, which is associated with development of intracranial aneurysms \citep{zhao2022study} and plaque instability in carotid arteries \citep{li2008assessment}.

Not all 3D vascular models are directly suitable for such downstream analysis tasks. For example, accurate curvature measurements require vessel centerlines \citep{kashyap2022accuracy}, and CFD requires a smooth representation of the vessel wall at a sub-voxel resolution \citep{kretschmer2013interactive}. Manual annotation of 3D vascular models meeting all these requirements from 3D image data is a laborious task and prone to inter- and intra-observer variability \citep{scherl2007semi}. Therefore, methods have been developed to automate the segmentation of vessels from 3D images \citep{lesage2009review, moccia2018blood}.

Over the last few years, deep learning-based methods for vessel segmentation have become popular \citep{litjens2017survey, chen2020deep}. Convolutional neural network (CNN) architectures, e.g. U-Net and nnU-Net \citep{ronneberger2015u,  cciccek20163d,isensee2021nnu} have achieved outstanding performance in segmenting vessels, such as aneurysms in the abdominal aorta \citep{lopez20193d}, liver vessels \citep{huang2018robust}, coronary arteries \citep{huang2018coronary} and retinal vessels \citep{fu2016deepvessel}. These CNN-based methods result in voxel mask representations of the vascular structures, which typically have high overlap with ground-truth segmentations, however, downstream tasks such as CFD require a 3D vascular model of sub-voxel resolution \citep{taylor2023patient}. Moreover, voxel masks may contain anatomical inconsistencies such as holes or disconnected parts, despite recent efforts towards topology-aware loss functions \citep{araujo2021topological, shit2021cldice}. Alternatively, topological guarantees can be included as an inductive bias by parametrising the model as a generalised cylinder \citep{shani1984splines}. The vessel is represented as a centerline and a set of contours locally orthogonal to this line. This approach has been adapted in previous automatic vessel segmentation methods \citep{lugauer2014improving, wolterink2019graph, alblasSPIE}. For such methods, reliable centerline extraction forms a crucial step. Additionally, centerlines are used to obtain a stretched visualisation of the vessel, which is used for, e.g., diameter measurements in abdominal aortic aneurysms \citep{manning2009abdominal}.

\begin{figure*}[t!]
    \centering
    \includegraphics[width=\textwidth]{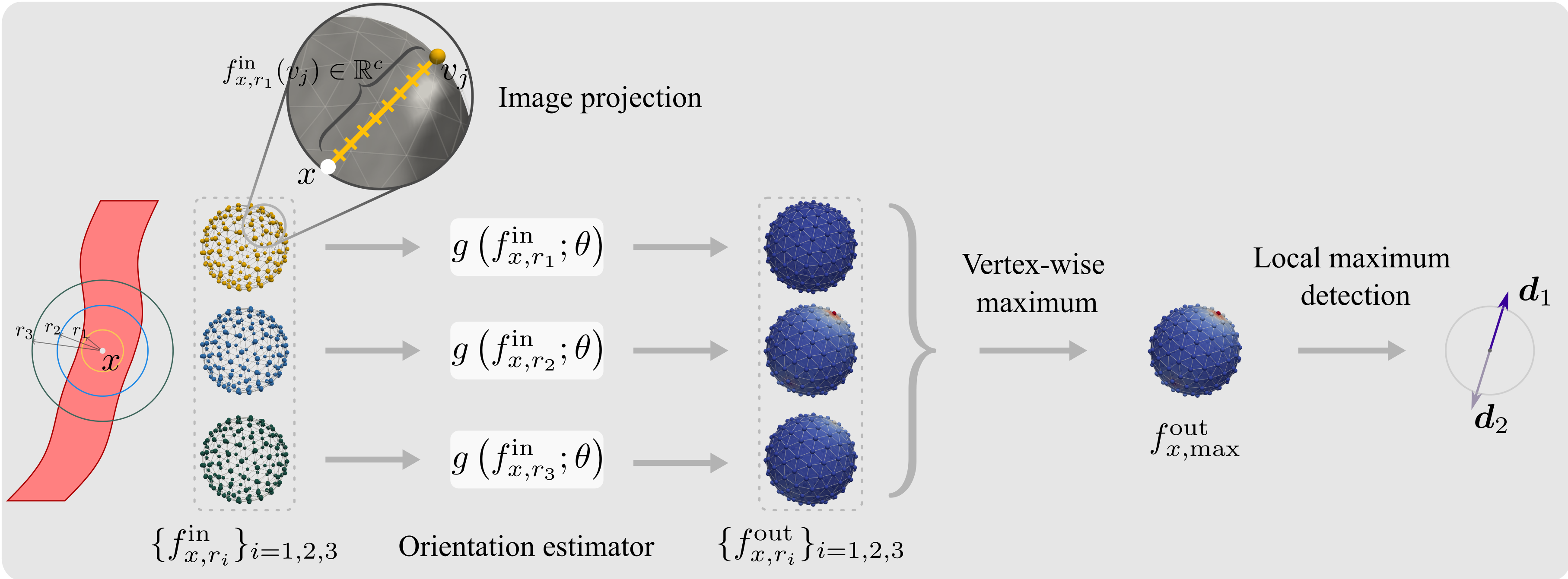}
    \caption{Schematic overview of our scale-invariant, rotation equivariant local vessel orientation estimator (SIRE). Local image information is extracted within a spherical volume and projected on the surface of the sphere at multiple scales ($r_1, r_2, r_3$). A graph convolutional network (GCN) $g(\cdot;\theta)$ with shared weights $\theta$ processes this information at each scale in parallel and obtains the local vessel orientation as a heatmap on the vertices of the discrete spherical domain. For each vertex, the maximum across the scales is obtained, forming the final prediction on a unit sphere, $f^{\text{out}}_{x, \text{max}}$. The local vessel orientations $\bm{d}_1$ and $\bm{d}_2$ are obtained by finding the two local maxima of $f^{\text{out}}_{x, \text{max}}$.}
    \label{fig:method}
\end{figure*}

Centerline extraction in tubular structures has a long history. A popular type of methods is open-curve snakes, where an energy functional consisting of internal and image-based forces is optimised from an initial seed point \citep{wang2011broadly, bekkers2015pde}. Another option is to find an optimal cost path between two or more points in the image subject to an image-based cost function \citep{li2007vessels}. These methods process global image information to find vessel centerlines, which can be costly in the case of high-resolution 3D volumes. Alternatively, iterative tracking-based approaches explore a volume locally, starting at a pre-defined seed point. Such algorithms iteratively determine the local vessel orientation, take a step forward, and terminate when some pre-defined or learned criterion is met. Determining the vessel orientation can be done using handcrafted features, e.g. Hessian eigenvalue analysis \citep{cetin2012vessel, kumar20133d} or template matching \citep{friman2010multiple}. More recently, trackers have emerged that use a CNN to locally determine the orientation of tubular structures such as coronaries \citep{wolterink2019coronary, gao2021joint, salahuddin2021multi}, vertebral arteries \citep{su2023deep} or even intestines \citep{van2022untangling}. The predictions of this CNN can then be integrated into a tracking algorithm using, e.g. single step \citep{wolterink2019coronary}, reinforcement learning \citep{su2023deep}, or multi-agent tracking \citep{van2022untangling}. The common denominator of such methods is that they employ a 3D CNN operating on local cubic image patches to estimate the vessel orientation.

Estimating the vessel orientation using 3D CNNs has two major limitations. First, arteries in the human body appear at many different calibres: coronary arteries that are distinguishable in typical MRI or CT data have diameters ranging between 1.9 and 4.5 mm \citep{dodge1992lumen}, whereas the diameter of the descending aorta ranges between 25 and 29 mm in healthy subjects \citep{erbel2001diagnosis}, and can reach 80 mm and above in patients diagnosed with an AAA \citep{aggarwal2011abdominal}. Hence, an orientation classifier trained using cubic patches of fixed size on a set of arteries with a specific calibre \citep{wolterink2019coronary, gao2021joint, su2023deep} will fail in the case of much larger or smaller arteries, as the network's fields-of-view are completely misaligned for arteries of different diameters. The application of such a classifier in arteries of a different calibre requires retraining, involving collection and manual annotation of new datasets; an extremely costly and time-consuming process. An orientation estimator that is unaffected by changes in the vessel's calibre, i.e. scale-invariant, is hence highly desired. Second, vessel tortuosity can vary widely per individual and among different anatomic regions in the human body. Due to this tortuosity, the vessel orientations in local image patches are not canonical. Previous CNN-based orientation estimators, however, operate on canonically oriented cubical patches and are \textit{not} equivariant to these changes in vessel orientation, i.e. SO(3)-equivariant. To deal with varying vessel orientations, previous works \citep{wolterink2019coronary, gao2021joint, su2023deep} have used rotation augmentation. However, data augmentation does not guarantee robustness to different orientations and might lead to undesirable increases in training time.

In this work, we present a modular local artery orientation estimator, with two key novelties: \textit{Scale-Invariance} and \textit{Rotation-Equivariance} (SIRE). We embed SIRE in an iterative tracking algorithm that sparsely traverses through a 3D volume starting from an automatically or manually determined seed point. SIRE's symmetry-preserving properties allow for estimating the local orientation of vessels of any calibre and tortuosity, without loss of performance. This generalisation can be leveraged during \textit{inference}, but also during \textit{training}. In this work, we demonstrate the generalisation of the orientation regressor using three diverse 3D CT angiography datasets containing vessels of various diameters and tortuosity: the public Vascular Model Repository (VMR), containing pulmonary arteries, coronary arteries, aortas and various branches of the aortofemoral tree \citep{wilson2013vascular}, the Automated Segmentation of Coronary Arteries (ASOCA) dataset \citep{gharleghi2023annotated}, including coronaries, and an in-house dataset with abdominal aortic aneurysms (AAAs). We perform numerical experiments demonstrating the rotation-equivariance and scale-invariance of SIRE. Finally, we demonstrate how SIRE generalises to determine the local orientation of arteries of any tortuosity and diameter.

\section{Methods}

\subsection{Orientation estimator}\label{sec:or_reg}
We consider the following local artery orientation estimation problem. Let $x \in \mathbb{R}^3$ be a position inside the artery lumen, from which we aim to infer local up- and downstream orientations $\bm{d}_1, \bm{d}_2 \in \mathbb{R}^3$ respectively. Note that due to vessel curvature, in general, $\bm{d}_1 \neq -\bm{d}_2$. We construct a local image patch $f_{x, r}^{\text{in}}$ of real-world size $r$ centered at $x$, that serves as input to an orientation estimator $g(\cdot; \theta): f^{\text{in}}_{x,r} \mapsto f^{\text{out}}_{x,r}$, with trainable weights $\theta$. From $f^{\text{out}}_{x,r}$, the local orientations $\bm{d}_1$ and $\bm{d}_2$ can be inferred. In existing trackers \citep{wolterink2019coronary, gao2021joint, su2023deep}, $g(\cdot; \theta)$ is a CNN, $f^{\text{in}}_{x,r}$ is a canonically oriented cubic image patch of fixed scale $r$, and $f^{\text{out}}_{x,r}$ is a probability distribution, whose local maxima correspond to $\bm{d}_1$ and $\bm{d}_2$.

We identified two limitations in existing orientation estimators. First, in existing methods, $f^{\text{in}}_{x,r}$ is only considered for a single scale $r$. Hence, local patches of, e.g., a coronary artery and the aorta will contain considerably different context. A patch that is suitable for estimating the orientation of a coronary artery will likely fall entirely into the aortic lumen, and conversely, a patch suitable for estimating the orientation of an aorta will likely not contain sufficient detail to estimate the orientation of a coronary artery. SIRE overcomes this issue by constructing a set of \textit{multi-scale} image patches: $\{f^{\text{in}}_{x, r_i} \}_{r_i \in R}$ on a set of $m$ predefined scales $R$, that are processed \textit{in parallel} by the estimator $g(\cdot; \theta)$, with shared weights $\theta$. This results in a set of $m$ scale-wise outputs $\{f^{\text{out}}_{x, r_i}\}_{r_i \in R}$ that are processed to a final prediction in a permutation invariant way, i.e. taking the maximum over all scales. Hence, this final prediction is an aggregated response of the spherical image patches across all scales in $R$ and is used to optimize $g(\cdot, \theta)$, without explicit restrictions on scale-wise responses. This encourages SIRE to independently learn meaningful features to determine an artery's orientation, regardless of its calibre. Scale symmetry is challenging to preserve, as it is a semi-group, meaning it is not closed, and invariance to a finite subset of group actions can be achieved. SIRE is invariant to scales within the range of the set $R$. Figure \ref{fig:method} shows an overview of our proposed method.

\begin{definition}[Equivariance]\label{def:equivariance}
    Let $\mathcal{X, Y}$ be Hilbert spaces, and let $G$ be a group with elements $g$. $g$ acts on $\mathcal{X}$ and $\mathcal{Y}$ through representations $\rho_\mathcal{X}$, $\rho_\mathcal{Y}$, respectively. A mapping $f: \mathcal{X} \to \mathcal{Y}$ is said to be \textit{equivariant} to $G$ if and only if $\rho_\mathcal{Y}(f(x)) = f(\rho_\mathcal{X}(x))$ $\forall x \in \mathcal{X}$, $\forall g \in G$.
\end{definition}

\begin{figure}[t!]
    \centering
    \includegraphics[width=\columnwidth]{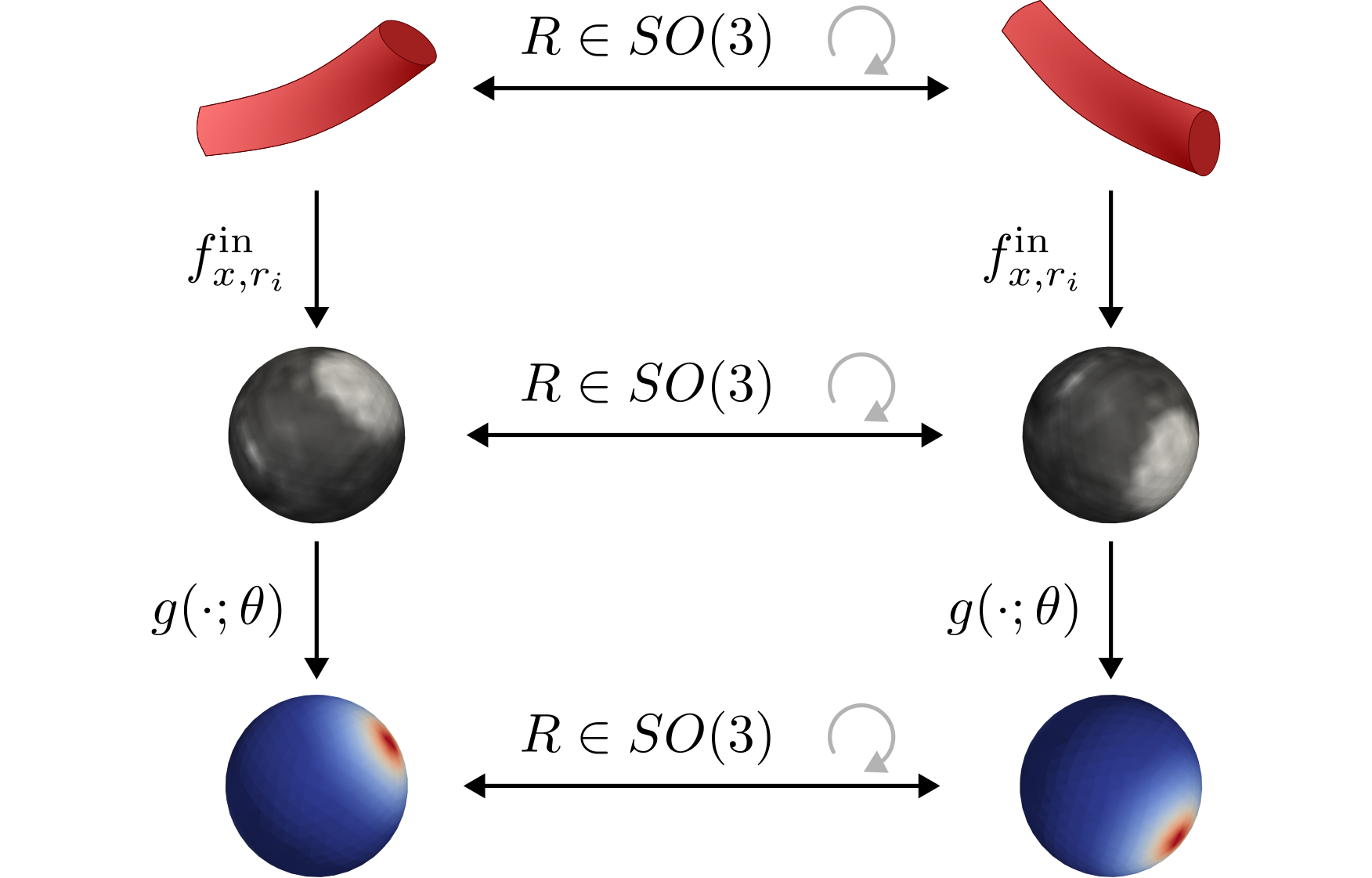}
    \caption{Commutative diagram of SO(3)-equivariance in SIRE. Rotation of the vessel orientation in the image data results in rotated features $f^{\text{in}}_{x, r_i}$. Processing with SO(3)-equivariant $g(\cdot;\theta)$ results in rotated $f^{\text{out}}_{x, r_i}$, conform Definition \ref{def:equivariance}.}
    \label{fig:commutative_diagram}
\end{figure}

A second key contribution of SIRE is rotation equivariance. Existing CNN-based orientation estimators operate on canonically oriented cubic image patches. However, due to vessel tortuosity, there is no canonical orientation of vessels on local image patches. A CNN is in general \textit{translation}-equivariant, but does not satisfy the equivariance property in Definition \ref{def:equivariance} for rotations $R \in SO(3)$. Instead, we choose to process image information on a \textit{spherical} domain, as shown in Figure \ref{fig:method}. By letting $g(\cdot; \theta)$ map from and to a spherical domain, the problem of vessel orientation estimation becomes completely intrinsic to the spherical domain and is independent of the orientation in ambient space. Transforming the data to a spherical domain poses one important challenge: a CNN-based architecture can no longer be used to process $f^{\text{in}}_{x,r}$, as such an architecture only works on Euclidean domains. Instead, we use a graph convolutional network (GCN), that aggregates information and makes a prediction on the spherical manifold, as shown in Figure \ref{fig:method}. This causes SIRE to be rotation-equivariant, in contrast to previously introduced CNN-based orientation classifiers. Figure \ref{fig:commutative_diagram} displays the rotation-equivariant behaviour of SIRE.

\subsubsection{Learning on spheres}\label{sec:input}
SIRE operates on images extracted on a spherical domain that we project to the surface of the unit sphere $S_1(0):=\{x \in \mathbb{R}^3 \hspace{1pt} : \hspace{1pt} ||x|| = 1\}$. This domain is discretised into a mesh $\mathcal{M}$, with $N$ equidistantly spaced vertices $\mathcal{V}$ and undirected edges $\mathcal{E}$, i.e. an icosphere. On $\mathcal{M}$, we can construct $c$-dimensional signals $f: \mathcal{V} \to \mathbb{R}^c$. The input of SIRE is a signal $f^{\text{in}}_{x, r}$ on $\mathcal{M}$, representing spherical image features centered at $x$ with a radius $r$, as shown in Figure \ref{fig:method}. To obtain this input signal, we cast rays with physical length $r$ in the direction of each of the vertices $v_j$ in $\mathcal{V}$ starting from the center $x$ and evaluate the image intensity at $c$ equidistant locations along this ray. Each ray projects the feature vector to its corresponding vertex, $f^{\text{in}}_{x,r}(v_j)$; this vector contains image information within a distance $r$ of $x$ in the direction of $v_j$. The set of all these vectors belonging to $v_j \in \mathcal{V}$ forms the input signal $f^{\text{in}}_{x,r}$ around $x$ at scale $r$.

We cast orientation estimation as a \textit{regression} problem, in contrast to previous work, where it was cast as a \textit{classification} problem \citep{wolterink2019coronary}. The desired output of SIRE is a scalar signal $f^{\text{out}}_{x, r}: \mathcal{V} \to \mathbb{R}$ that represents the vessel orientations $\bm{d}_1, \bm{d}_2$ through its local maxima on the spherical surface. We adapt the method presented in \cite{sironi2015multiscale} to construct a target output signal $f^{\text{out}}_{x, GT}$ of $g(\cdot; \theta)$, where the response value for each vertex $v_i \in \mathcal{V}$ is based on its proximity to the normalised ground-truth $\bm{d}_1, \bm{d}_2$ at $x$:
\begin{align}\label{Eq: Gaussian_sphere}
    &f^{\text{out}}_{x, GT}(v_i) = \begin{cases}
    e^{\alpha\left(1 - \frac{D(v_i, \tilde{v})}{\beta} \right)} \hspace{0.5 cm} & \text{if } D(v_i, \tilde{v}) < \beta \\    
    0 \hspace{1 cm} &\text{otherwise},
    \end{cases}\\
    &\text{with } D(v_i, \tilde{v}) = \min\limits_{i \in \{ 1, 2 \}} ||v_i - \tilde{v}_i||_\mathcal{H} \nonumber \\
    & \text{and } \tilde{v}_i = \argmin\limits_{v\in \mathcal{V}} ||v - \bm{d}_i||_\mathcal{H}.
\end{align}
Here, $||\cdot||_\mathcal{H}$ is the Haversine distance over the surface of the sphere, $\tilde{v}_i$ is the vertex in $\mathcal{V}$ closest to the objective direction $\bm{d}_i$, $\beta$ is a predefined nonzero radius, and $\alpha$ is a control parameter. 

\subsubsection{Gauge-equivariant mesh convolution}
At the core of SIRE is a graph convolutional network (GCN), that processes the signals $f^{\text{in}}_{x,r}$ and predicts scalar-valued outputs: $f^{\text{out}}_{x,r}: \mathcal{V} \to \mathbb{R}$. This GCN consists of convolution layers, which aggregate information over the surface of $\mathcal{M}$. Convolution layers are defined through message passing \cite{gilmer2017neural}.
\begin{align}
    f^{k+1}(v_i) = \sigma \left(\phi \ast f^k(v_i) \right) = \sigma \left(\sum\limits_{v_j \in \mathcal{N}(v_i) \cup v_i} \phi f^k(v_j) \right),
\end{align}
where $\mathcal{N}(v_i)$ represents the neighbourhood of vertex $v_i$, $\sigma$ is a nonlinear activation function computed over the updated features and $\phi$ is a $C^{k+1} \times C^{k}$ matrix with trainable weights $\theta$. Depending on the kernel $\phi$, message passing is either isotropic or anisotropic. In isotropic message passing, $\phi$ is identical for all neighbouring vertices, meaning it yields the same result regardless of the features or relative positions of vertices. In anisotropic message passing, $\phi$ can depend on features of each of the neighbouring vertices, an example is graph attention (GAT) \citep{velivckovicgraph,brody2021attentive}, in which learned attention coefficients are computed:
\begin{align}\label{eq:GATconv}
    \sigma \left(\phi \ast_{\text{GAT}} f^k(v_i)\right) = \sigma \left( \sum\limits_{v_j \in \mathcal{N}(v_i) \cup v_i} \phi\left(f^k(v_i), f^k(v_j)\right) f^k(v_j) \right).
\end{align}

Note that this form of anisotropic message passing does \textit{not} take the geometry of the manifold into account, i.e. the relative positions of each of the vertices $v_i \in \mathcal{V}$.

The main challenge in distinguishing neighbouring vertices based on relative positions is the fact that there is no canonical orientation on a manifold and hence no unique way to define the orientation of the kernel $\phi$. Positions of neighbouring vertices $v_j \in \mathcal{N}(v_i)$ can be described by a polar coordinate system at the tangent plane of $v_i$, after projecting the neighbouring vertices on this tangent plane, using one neighbour as a reference. The choice of this neighbour, also called gauge, is arbitrary and choosing a different neighbour should not affect the computed message.

In gauge-equivariant mesh (GEM) convolution, $\phi$ is constructed to be equivariant to these gauge changes, i.e. group actions from SO(2). To be equivariant under transformations in this group, $\phi$ is constrained to a lower-dimensional subspace \citep{cohen2016group}, and signals $f^{\text{in}}$ are written in terms of a linear combination of the irreducible representations of SO(2). 

Before messages between vertices can be computed, their features should be expressed in the same vector space. We use a parallel transporter $\rho_{v_j\to v_i}$, that uniquely transforms features defined at $v_j\in\mathcal{N}(v_i)$ to the tangent plane of $v_i$. Together with $\phi$ they form the two ingredients of GEM convolution:

\begin{align}\label{eq:GEMconv}
     \sigma \left(\phi \ast_{\text{GEM}} f^k(v_i) \right) = \sigma \left(\sum\limits_{v_j \in \mathcal{N}_{v_i}\cup v_i} \phi(\theta_{v_i, v_j}) \rho_{v_j \to v_i} f^k(v_j)) \right).
\end{align}
Here, $\phi$ is conditioned on the angle $\theta$ between vertex $v_i$ and $v_j \in \mathcal{N}(v_i)$ in the tangent space of $v_i$, enabling the filter to distinguish neighbours based on their relative angles. For a detailed derivation of $\phi$, we refer the reader to \cite{dehaan2021}.

We used GCNs with either graph attention convolutions (Eq.\ref{eq:GATconv}) or GEM convolutions (Eq.\ref{eq:GEMconv}) to process our image data on $\mathcal{M}$ and predict the scalar output signal $f^{\text{out}}$. As the convolutions in both networks are intrinsic, and the in- and outputs of the network are defined on $\mathcal{M}$, our orientation estimation method is rotation-equivariant. The difference between the two networks lies in their kernel expressiveness. In GAT convolutions, vertices are distinguished through an attention mechanism, whereas the GEM convolutions have a sense of direction.

\begin{figure}[t!]
    \centering
    \includegraphics[width=\columnwidth]{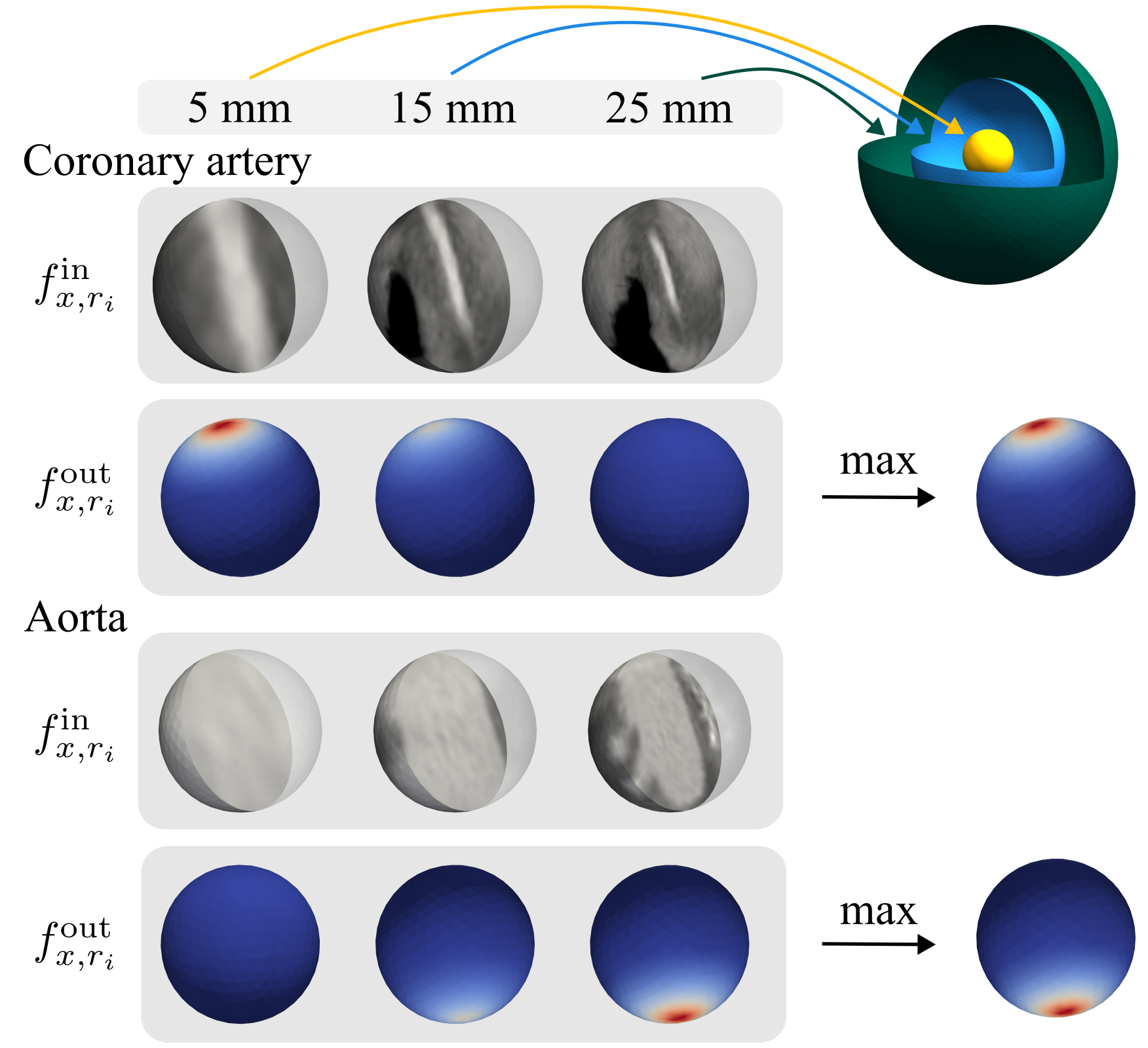}
    \caption{Spherical input features $\{f^{\text{in}}_{x, r_i}\}_{r_i \in \{5, 15, 25\}}$ for a coronary artery and aorta and their $\{ f^{\text{out}}_{x, r_i} \}_{r_i \in \{ 5, 15, 25 \}}$. The response for the coronary at a 5 mm scale is similar to the response for the aorta at a 25 mm scale.}
    \label{fig:scales}
\end{figure}

\subsubsection{Scale invariance}

To let the orientation estimator generalise across vessels of different calibres, we consider the spherical image signals with $m$ different radii $R = \{ r_1, r_2, ..., r_m \} \subset \mathbb{R}$, as shown in Figure \ref{fig:method}. We embed spherical image data with radii $r_i \in R$ into the icosphere $\mathcal{M}$ as described in Section \ref{sec:input}. This results in $m$ input signals $f^{\text{in}}_{x, r_i}$ at each scale in $R$, each defined on the \textit{same} set of vertices $\mathcal{V}$. The vertex-wise signals at each scale are considered to be aligned and correspond to the \textit{same} direction. Note that the amount of context and detail in $f^{\text{in}}_{x, r_i}$ differs per scale: namely, if $r_1 < r_2$, $f^{\text{in}}_{x, r_1}(v_i)$ contains less context information but at a higher level of detail than $f^{\text{in}}_{x, r_2}(v_i)$. More importantly, the input signals of large vessels at large scales look similar to the input signals of small vessels at small scales. This effect is shown in Figure \ref{fig:scales}, where the spherical image signal of a coronary artery (radius 2 mm) at a scale of 5 mm looks similar to that of the aorta (radius 15 mm) at a 25 mm scale. 

To make SIRE scale-invariant, we process the input signal at multiple scales through the GCN $g\left(\cdot; \theta \right)$ \textit{in parallel}, as depicted in Figure \ref{fig:method}. This results in $m$ independent scalar output signals, \{$f^{\text{out}}_{x, r_i}\}_{r_i \in R}$ on $\mathcal{M}$, where the weights $\theta$ are shared among the scales. The final output of SIRE is $f^{\text{out}}_{x, \text{max}}$, which is obtained by taking the vertex-wise maximum of $f^{\text{out}}_{x, r_i}$ across $R$. Because information in different scales is combined using a maximum operator, there is a lot of freedom in choosing the scales during training: they may differ for each sample in the training set, as long as their fields of view contain a sufficient amount of context around the vessels present in the data. Even the number of scales considered during each forward pass through the network can change. Moreover, the ordering of scales is irrelevant, as the maximum operation used in the process is permutation invariant. After training, the user can provide a set of scales $R$ for the vessels of interest in new image data, possibly unseen during training, \textit{without} the need for additional retraining.

\subsubsection{Training strategy}\label{sec:train_strat}
Training SIRE requires a dataset of 3D images with manually annotated vessel centerlines $\gamma(t), \xspace t\in[0,\ell]$. Before training, we define a set of scales $R$ that will be considered. To generate one training sample, we randomly sample $\tilde{t} \in [0,\ell]$ from a uniform distribution and obtain a point $x=\gamma(\tilde{t}) \in \mathbb{R}^3$ on the centerline. Next, we construct the set of multi-scale spherical input features $\{f^{\text{in}}_{x, r_i}\}_{r_i \in R}$. If available in the training data, we use the vessel radius at $x$, $\rho(\tilde{t})$, to find the $\bm{d}_1, \bm{d}_2$ at $x$ using finite differences at $\gamma(\tilde{t} \pm \eta \rho(\tilde{t)})$ and normalise them. From $\bm{d}_1$ and $\bm{d}_2$ we determine the target response $f^{\text{out}}_{x, \text{GT}}$ (Sec. \ref{sec:input}).

During each forward pass through the network, we process the input features at different scales in parallel, resulting in $m$ scale-wise predictions with a scalar value on each vertex. As shown in Figure \ref{fig:method}, we take the maximum for each vertex across $R$. All these operations are differentiable, and therefore the final output of SIRE can be used directly to compute the loss with respect to $f^{\text{out}}_{x, \text{GT}}$ and update the weights of $g(\cdot; \theta)$. Any loss function for regression could be used, we achieved the fastest convergence with a mean squared error loss function. The network learns in a weakly supervised manner which scales contain the most useful information to determine local vessel orientation, \textit{without} being given any explicit instructions. Scales containing useful information will have higher scalar activations on $f^{\text{out}}_{x, r_i}$ than less useful scales and are hence aggregated into the final prediction, while scales that are too small or too large are ignored.

Radii of vessels in the training data are continuous, whereas the scales considered during training in SIRE form a discrete set. This means that vessels with a similar but unequal diameter may have the highest activation on one of the scales seen during training, and a true scale-invariant representation may not be learned. To become more robust against variations in vessel calibre, we train on \textit{randomized} scales in some of our experiments. Instead of picking a fixed set $R$ before training, $m$ scales are randomly sampled from a distribution $P$ on the fly during training. $P$ can be any probability distribution, we used a uniform distribution $\mathcal{U}_{[a, b]}$, whose boundaries depend on the radii of vessels in the dataset.

To encourage low activation values when none of the scales contain useful image information to determine the vessel direction, we also create \textit{negative} samples. With probability $0.1$, we sample a point near the vessel, outside the lumen. Again, the GCN processes the local multi-scale inputs. Instead of comparing the output to $f^{\text{out}}_{x, \text{GT}}$, we minimize the activations on $f^{\text{out}}_{x, \text{max}}$. This essentially creates a data-driven guardrail during tracking, right outside the vessel lumen.

\begin{figure}[t!]
    \centering
    \includegraphics[scale=1]{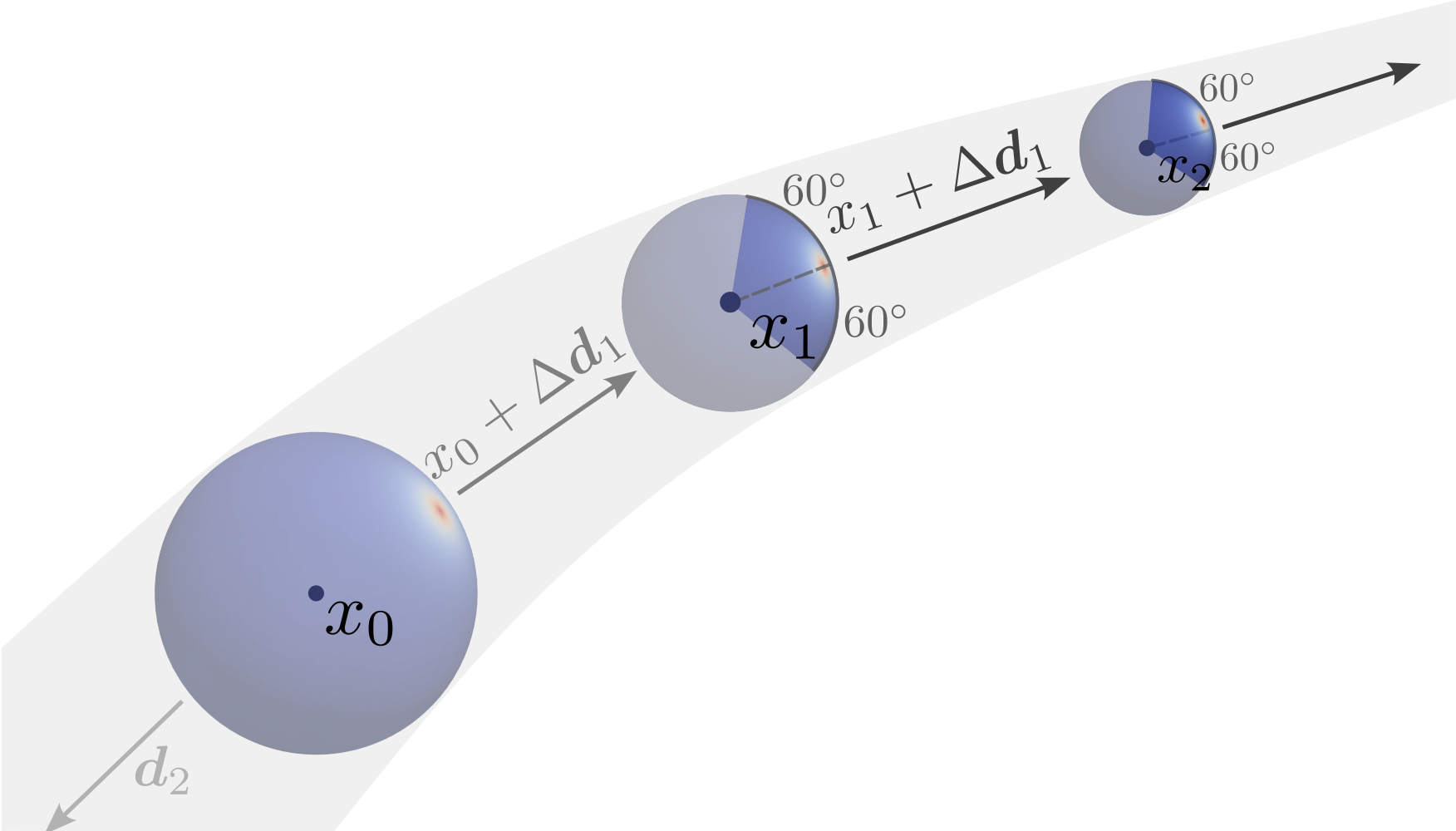}
    \caption{Tracking algorithm on a vessel with a decreasing radius. The tracker is initialized at $x_0$, where SIRE predicts local vessel orientations, adapting to the local vessel size, and traverses through the vessel with a step size $\Delta$ in direction $\bm{d}_1$ until a stopping criterion is met.}
    \label{fig:tracking_algorithm}
\end{figure}

\subsection{Iterative tracking}\label{sec:tracking}
The task that SIRE tackles is intentionally general: at any continuous point on or near a vessel centerline, it provides the local direction of that centerline. Vessel centerlines can be extracted by iteratively determining the local vessel direction. Before initialising the tracker, the scale set $R$ to consider is defined. Note that these scales do \textit{not} have to be the same as during training. More or fewer scales can be considered, as long as they align with the radius of the vessels present in the image. Moreover, these scales could change for each step of the iterative tracking.

After defining $R$, a seed point $x_0 \in \mathbb{R}^3$ is passed to initiate the tracking procedure. Seed points can be defined manually by a user, or automatically based on an initial estimate of the location of the arteries, obtained using, e.g., a segmentation \citep{ ronneberger2015u,isensee2021nnu}. Once an initial seed point $x_0$ is set, we construct the multi-scale spherical inputs $\{f^{\text{in}}_{x_0, r_i}\}_{r_i \in R}$. Again, we take the maximum across $m$ scales to obtain $f^{\text{out}}_{x_0, \text{max}}$. In the first step of tracking, there is no moving direction from a previous iteration. Hence, both directions $\bm{d}_1$ and $\bm{d}_2$ are considered and added to a queue. The first objective direction is found by taking the argmax of $f^{\text{out}}_{x_0, \text{max}}$, subsequently we take the argmax again, after masking out a $ 90^{\circ}$ region around $\bm{d}_1$.

We first traverse through the vessel in the direction of $\bm{d}_1$ with a fixed step size $\Delta$. In the next step, we construct a new multi-scale spherical input at $x_1 = x_0 + \Delta \cdot \bm{d}_1$. Again, we obtain $f^{\text{out}}_{x_1, \text{max}}$ from the GCN that can be used to obtain the vessel directions at $x_1$. To prevent the tracker from reversing, the vessel direction at $x_1$ is determined by finding the local maximum of $f^{\text{out}}_{x_1,\text{max}}$ within $60^{\circ}$ of $\bm{d}_1$ \citep{wolterink2019coronary} (Figure \ref{fig:tracking_algorithm}).

This iterative process repeats until a stopping criterion is met. This criterion is based on an uncertainty metric on the output of $g(\cdot; \theta)$. We define uncertainty as the entropy of $f^{\text{out}}_{x, \text{max}}$ after transforming it to a probability distribution using a softmax function. Once the entropy exceeds a threshold value of $\tau$, the tracking procedure terminates. High entropy implies that the tracker cannot find any useful information within the provided scales and has likely left the vessel lumen or continued tracking the vessel until it became too small to distinguish. However, diseased regions in the arteries, e.g. stenosis and calcifications may also result in a local high entropy. To encourage the tracker to continue tracking through these regions, we take a moving average of the entropy over the last five steps. If direction $\bm{d}_2$ has not yet been explored, the tracking algorithm is re-initialised at $x_0$ and tracks the vessel in the direction of $\bm{d}_2$. Otherwise, the vessel is considered fully tracked and the algorithm terminates.

\subsection{Automatic vessel tree extraction}\label{sec:vesseltree}
The tracking algorithm described above relies on a seed point that is placed inside the vessel lumen. These seed points can be manually placed by a user, or found automatically. Seed points are put in a queue, and we initialise the tracking algorithm at an arbitrary point from this queue. After the termination of the tracker, all the visited points are removed from the queue. This process continues until the queue is fully empty, resulting in a tracked vascular tree structure.

Note that the automatic segmentation method needed to obtain the queue of seed points requires retraining when applied to datasets containing other vessels, in contrast to SIRE.

\begin{figure}[t!]
    \centering
    \includegraphics[scale=1]{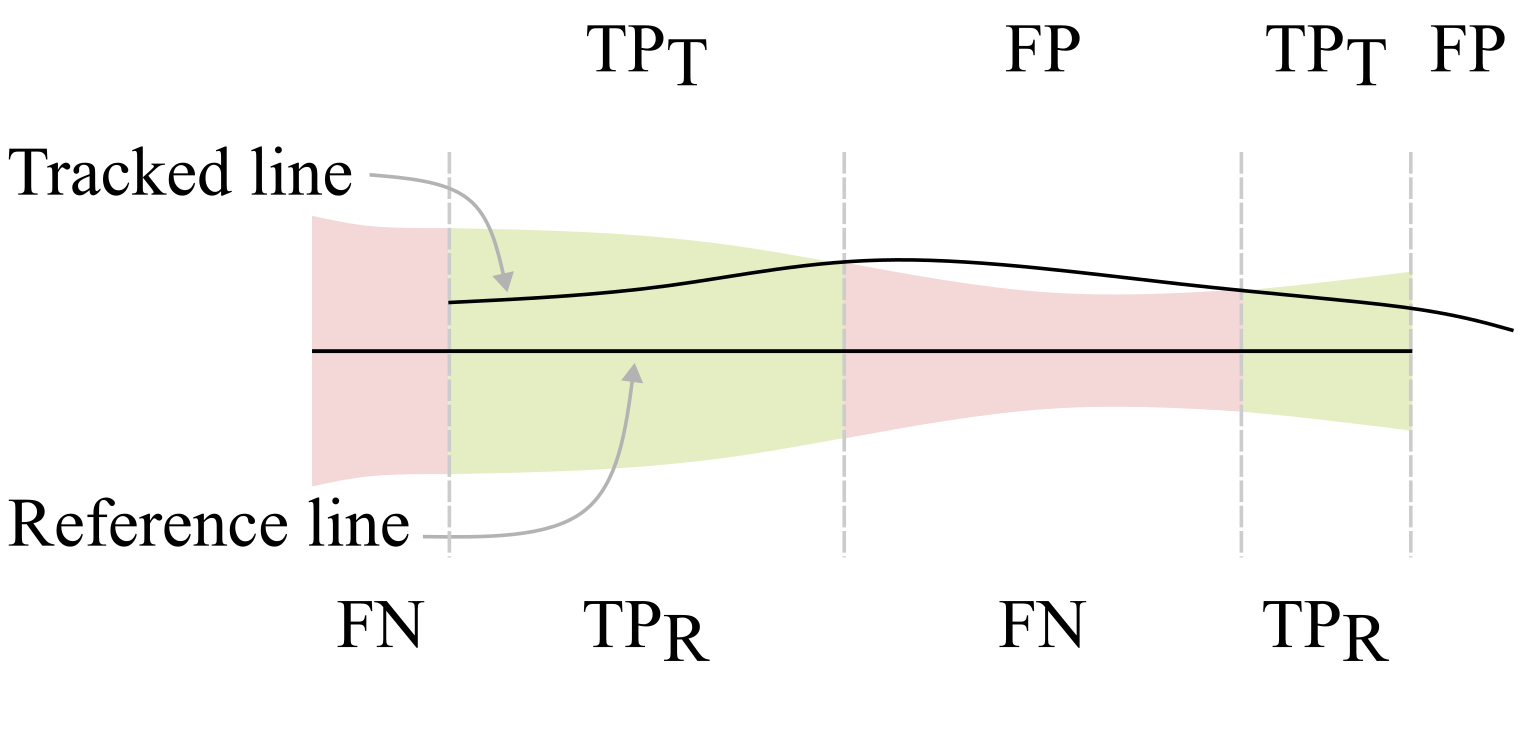}
    \caption{Evaluation metrics for centerline tracking. Points on the reference and tracked line are labelled as true positive, false positive or false negative, based on the local radius of the vessel. Image adapted from \cite{schaap2009standardized}.}
    \label{fig:metrics}
\end{figure}

\subsection{Evaluation metrics}
To quantify the performance of the vessel orientation estimations, we use the cosine similarity metric. We compute this metric between the ground-truth vessel orientations $\bm{d}_1$ and $\bm{d}_2$ at location $x$, and the vessel orientations predicted by SIRE by finding the two local maxima from $f^{\text{out}}_{x, \text{max}}$ at least 60\textsuperscript{$\circ$} apart. The cosine similarity ranges between -1 and 1, where 1 indicates perfect alignment between the directions, 0 implies orthogonality, and -1 means the directions are opposite.

To quantitatively evaluate the quality of tracked centerlines, we adapt the metrics introduced in \cite{schaap2009standardized}. The tracked and reference centerlines are compared by connecting the points on each of the lines to the closest point on the other line. Figure \ref{fig:metrics} shows how points on both lines are classified as true positive, false positive or false negative, if the ground-truth radius of the artery is available along the reference line, based on the proximity of both lines. On the \textit{tracked line}, points are classified as TP\textsubscript{T} if the closest point on the reference centerline lies within the local ground-truth radius of the vessel, or FP if the closest centerline point is further away. Similarly, the points on the \textit{reference line} are classified as TP\textsubscript{R} if there is a point on the tracked line within the vessel radius, or FN otherwise. Using these classes, we can determine precision, recall and the F1 score, which are essentially the same metrics as introduced in \cite{shit2021cldice}.
\begin{itemize}
    \item \textbf{Precision}: Fraction of the \textit{tracked} line that lies within a radius distance of the reference line. $\frac{|\text{TP\textsubscript{T}}|}{|\text{TP\textsubscript{T}}| + |\text{FP}|}$. 
    \item \textbf{Recall}: Fraction of the \textit{reference} line that lies within a radius distance of the tracked line. $\frac{|\text{TP\textsubscript{R}}|}{|\text{TP\textsubscript{R}}| + |\text{FN}|}$
    \item \textbf{Overlap}: Combination between the precision and recall, also known as the F1 score and equivalent to the Dice similarity coefficient for centerlines \citep{shit2021cldice}:
    $\frac{|\text{TP\textsubscript{T}}| + |\text{TP\textsubscript{R}}|}{|\text{TP\textsubscript{T}}| + |\text{TP\textsubscript{R}}| + |\text{FP}| + |\text{FN}|}$
    \item \textbf{Average inside (AI)}: average distance between the \textit{tracked} line and \textit{reference} line for the TP\textsubscript{T} points. AI is an accuracy metric that is independent of the length of the tracked centerline.
\end{itemize}

\begin{figure}[t!]
    \centering
    \includegraphics[width=\columnwidth]{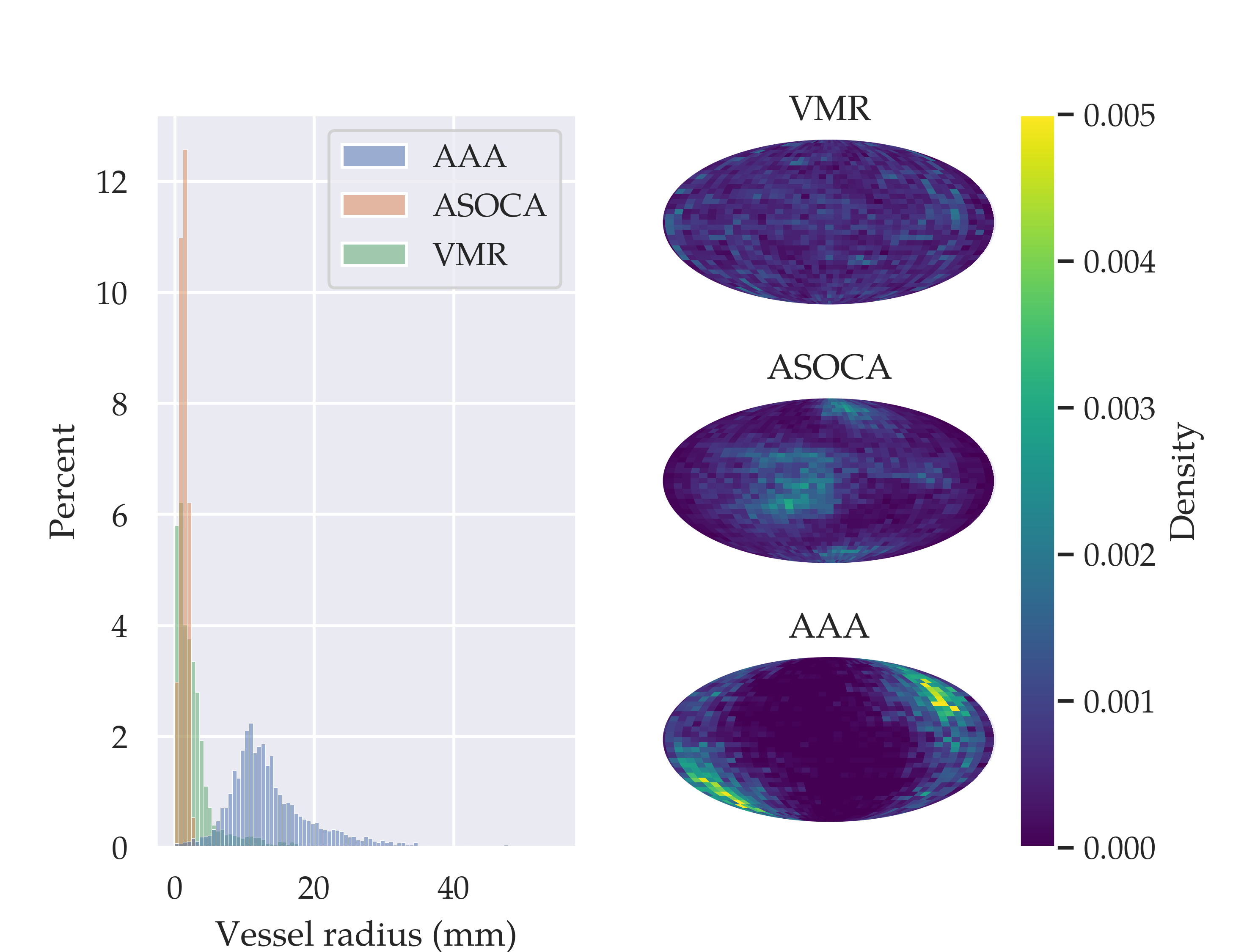}
    \caption{Overview of the radii and objective directions in the VMR, ASOCA and AAA datasets for 10000 randomly drawn samples from the annotated centerlines. \textit{Left:} Histogram of the vessel radii for the VMR, ASOCA and AAA datasets, showing that the ASOCA dataset has very homogeneous vessel radii, in contrast to the AAA and VMR datasets. The AAA dataset contains vessels with much larger radii. \textit{Right:} Mollweide projection of the densities of objective directions $\bm{d}_1, \bm{d}_2$, indicating that the vessel orientations in VMR are very heterogeneous, ASOCA are focused around left- and right coronary arteries and orientations in the AAA dataset are very homogeneous.}
    \label{fig:vis_dataset}
\end{figure}

\section{Data}\label{sec:data}
SIRE should generalise to vessels of varying sizes and tortuosity due to its scale-invariant and rotation equivariant design. This property can be leveraged both during \textit{training} and during \textit{inference}. Therefore, datasets containing a diverse range of vessel calibres with accurate centerline annotations and local radii would be ideal for training SIRE. To demonstrate the generalisation of SIRE, we used three different vessel datasets containing different ranges of vessel calibres and vessel tortuosities, as shown in Figure \ref{fig:vis_dataset}.

The first dataset is the Vascular Model Repository (VMR) \citep{wilson2013vascular}; a publicly available, very comprehensive dataset containing vessels in different anatomical regions. Second, we used the publicly available Automated Segmentation of Coronary Arteries (ASOCA) dataset \citep{gharleghi2023annotated}, containing coronary computed tomography angiography (CCTA) images, accompanied by centerlines and segmentation masks of the full coronary tree. Third, we used an in-house dataset of CTA scans of patients with abdominal aortic aneurysms, on which the centerlines and lumen contours were annotated in the abdominal aorta.  These three datasets have different compositions in terms of vessel radii and orientations, as shown in Figure \ref{fig:vis_dataset}. The arteries included in the VMR and AAA datasets have a wider varying calibre than the arteries in the ASOCA dataset. As expected, the median radius of the arteries in the AAA dataset is much larger than that of the other two datasets. Moreover, Figure \ref{fig:vis_dataset} shows that the AAA and ASOCA datasets have clear hot spots in vessel orientation. In contrast, vessel orientations in the VMR dataset are more dispersed.

\subsection{Vascular Model Repository}
The Vascular Model Repository (VMR) \citep{wilson2013vascular} is a publicly available database containing 3D vascular models of 206 human and animal subjects. This dataset was built to develop and validate blood flow simulation methods. The repository is divided into vascular models of five different anatomical regions: aortofemoral tree, aortic arch and thoracic aorta, coronaries, pulmonaries and vertebral arteries, all including vessels of widely varying diameters. Most vascular models include MR or CT data with vessel annotations. These annotations are tubular parametrisations \citep{shani1984splines} and consist of vessel centerlines and contours drawn orthogonally to the vessel centerline, that were made using the SimVascular software \citep{updegrove2017simvascular}. The VMR includes both male and female subjects, with ages ranging between infants to 79 years old. Moreover, the database contains healthy and diseased subjects in each of the five anatomical regions, e.g. abdominal aortic aneurysms, aortic dissections or coronary artery disease. The arteries of some subjects may contain stents.

For the experiments in this work, we selected the human subjects from the VMR that were accompanied by a CTA scan, as well as a 3D model of the vascular structure. We used the 3D vascular models to estimate the local vessel radius along the centerlines, to determine the reference directions along the centerlines during training. In total, we included 41 subjects, from which 16 were in the aortofemoral region, 7 in the aortic region and 18 in the coronary region. As the contrast levels in the pulmonary arteries were much lower than in the coronary and aortic region, we decided to leave the pulmonary vessels out of the training data. The diversity of vessel radius and orientations in this dataset makes it very suitable to demonstrate the generalisation of SIRE. However, as the VMR was not designed for vessel tracking, the quality of the centerlines may be sub-optimal. We will therefore not assess centerline tracking performance on the VMR dataset.

\begin{figure}[t!]
    \centering
    \includegraphics[width=\columnwidth]
    {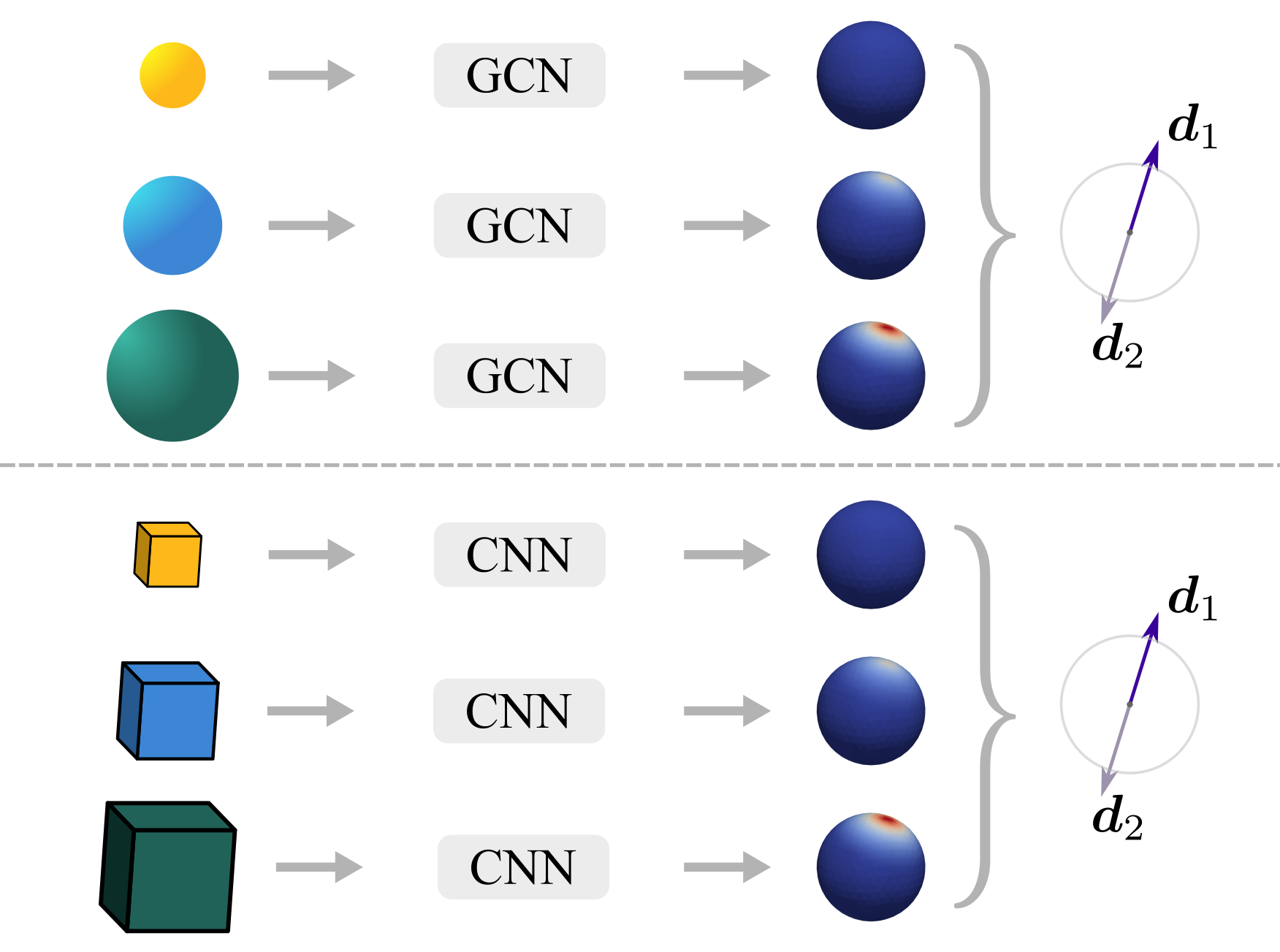}
    \caption{Schematic description of swapping out the GCN for a CNN in vessel orientation estimation. \textit{Top:} GCN processes multi-scale spherical inputs to predict scale-wise activations that are used to find the local vessel orientations. \textit{Bottom}: CNN processes multi-scale, canonically oriented, cubic inputs to predict scale-wise activations that are used to find the local vessel orientations.}
    \label{fig:exp_rot_eq}
\end{figure}

\subsection{ASOCA dataset}
The Automated Segmentation of Coronary Arteries (ASOCA) dataset consists of 60 coronary computed tomography angiography (CCTA) images. This set was released as part of a public challenge \citep{gharleghi2023annotated}. The images are anisotropic, with a slice thickness of 0.625 mm, and an in-plane resolution varying between 0.3 and 0.4 mm. The ASOCA dataset contains 30 healthy subjects, and 30 patients diagnosed with coronary disease. For both cohorts, a training set of 20 images with ground-truth segmentation masks, centerlines and local vessel radius are available for all coronaries with a diameter larger than 1 millimeter. Ground-truth segmentations of the remaining 10 images are kept by the challenge organizers as an independent test set. We used the 40 publicly available training subjects in the ASOCA dataset in our experiments.

\subsection{In-house Abdominal Aortic Aneurysm dataset}
The third dataset we use in this work is an in-house dataset consisting of 108 CTA scans of patients with an abdominal aortic aneurysm (AAA). All patients were treated with an EVAR procedure at Amsterdam UMC locations AMC and VUmc and were retrospectively included in our dataset. All scans contain at least the thoracic region until the iliac bifurcation and were made during the arterial phase, meaning that the contrast agent is present in the arteries. The slice thickness of these images ranged between 0.5 and 2.0 mm, while the in-plane resolution was generally higher and ranged between 0.625 and 0.98 mm.

In these CTA scans, centerlines and locally orthogonal contours of the abdominal part of the aorta were manually annotated between the top of the T12 vertebra and the iliac bifurcation. The centerlines were annotated by three independent observers, and the final centerline used for training was obtained by averaging between these three lines \citep{van2008averaging}. The locally orthogonal contours delineating the vessel lumen and thrombus were annotated separately by a single observer, and drawn every 10 mm along the centerline. These contours were used to obtain a watertight implicit representation of the surface of the vessel wall \citep{alblas2023going}, which were used to estimate the radius of the lumen along the centerline.

\begin{figure}[t!]
    \includegraphics[width=\columnwidth]{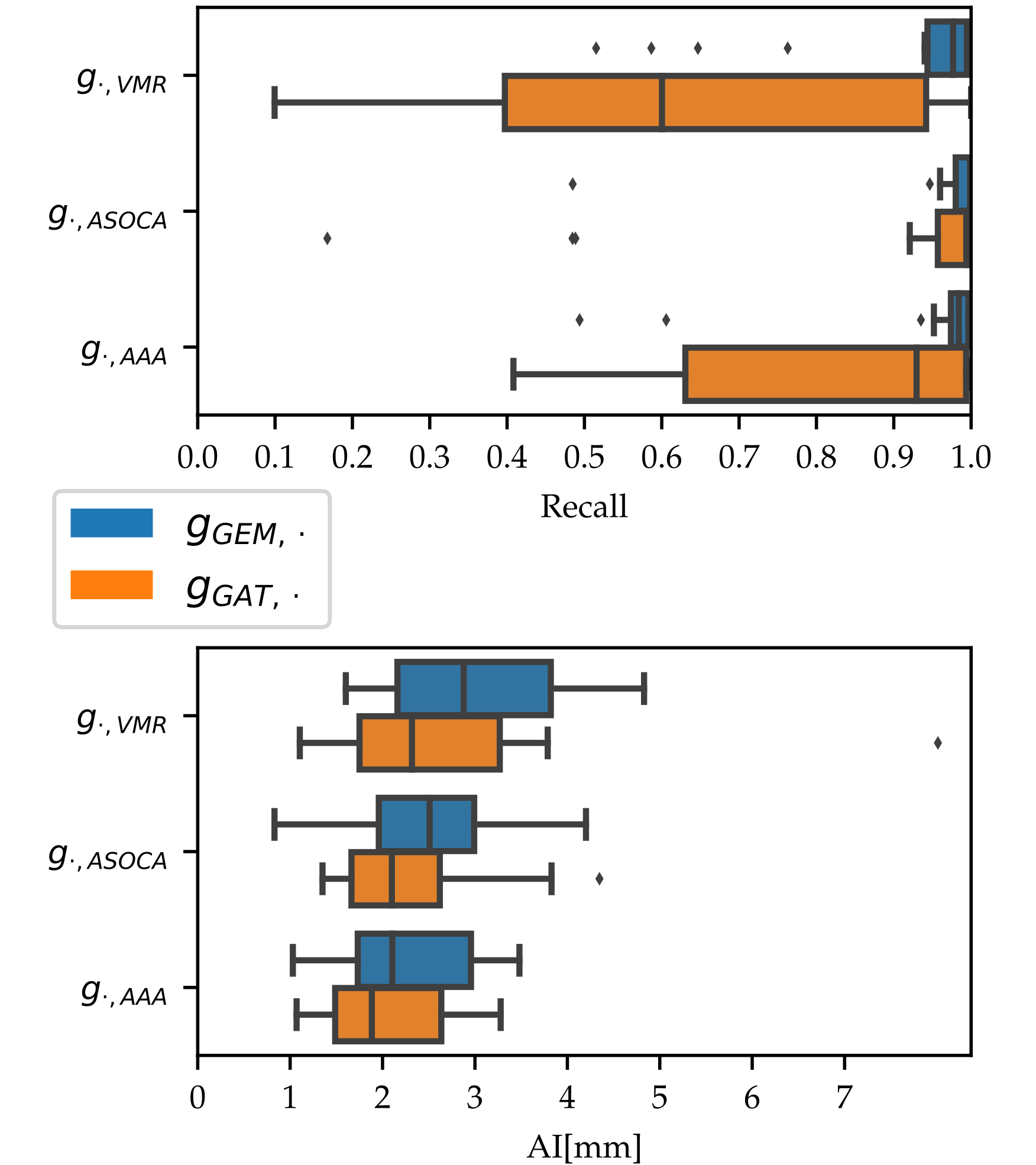}
    \caption{Quantitative metrics for the trackers $g_{ \text{GEM}, \{\text{VMR, ASOCA, AAA} \}}$ and $g_{\text{GAT},\{\text{VMR, ASOCA, AAA}\}}$ for tracking AAAs in AAA\textsubscript{test}. \textit{Top:} Recall values compared to the ground-truth centerline, \textit{Bottom:} Average inside distances between the tracked lines and the ground-truth centerlines.}
    \label{fig:AAA_AI}
\end{figure}

\begin{figure*}[t!]
    \centering
    \includegraphics[scale=1]{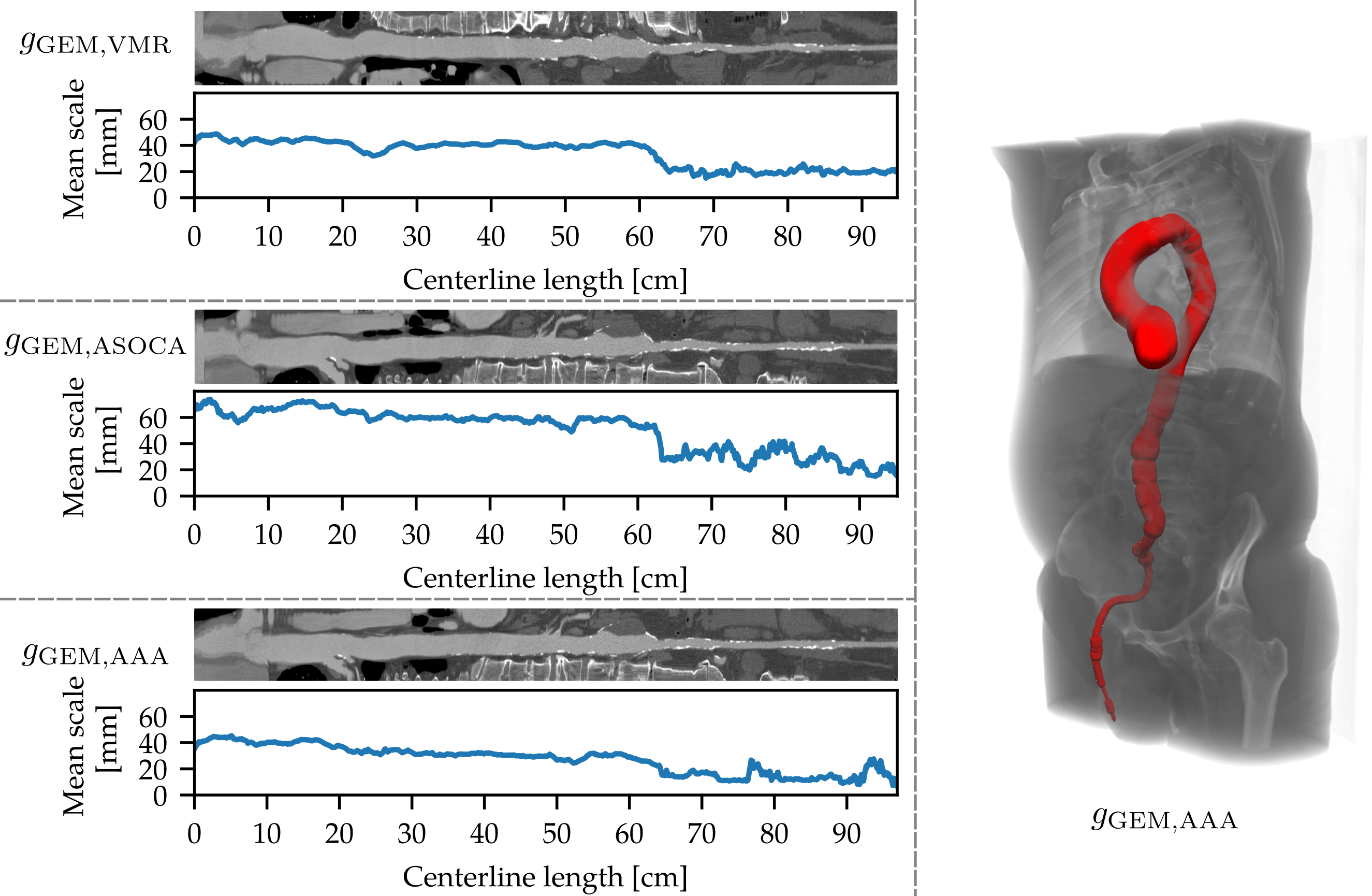}
    \caption{\textit{Left:} Stretched multi-planar reconstructions for the trackers $g_{\text{GEM}, \{ \text{VMR, ASOCA, AAA}\}}$ with the average active scale for one of the patients in AAA\textsubscript{test}. \textit{Right:} 3D rendering of the tracked centerline from $g_{\text{GEM, AAA}}$, starting from the aortic root, continuing through the aortic arch, abdominal aorta, iliac artery until the image boundary from a single seed point. Sphere radii indicate the average active scale.}
    \label{fig:AAA_MPR}
\end{figure*}

\section{Experiments and Results}\label{sec:exp}
We implemented the orientation estimator and the tracking algorithm in PyTorch. The trainable part was swapped out for three different architectures: a 3D CNN, a GCN with graph attention convolution \citep{brody2021attentive}, and a GCN with GEM convolution \citep{dehaan2021}, that we denote as $g_{\{\text{CNN, GAT, GEM}\}}$ respectively. To assess the importance of rotation equivariance in our method, we compare the rotation-equivariant, scale-invariant $g_{\{\text{GAT, GEM}\}}$ to the scale-invariant $g_{\text{CNN}}$ (Fig. \ref{fig:exp_rot_eq}). $g_{\text{CNN}}$ was trained on canonically oriented, cubic patches extracted at multiple scales, with resolution $63\times 63 \times 63$. In contrast, $g_{\{\text{GAT, GEM}\}}$ were trained on spherical input samples defined on the icosphere $\mathcal{M}$, which was discretised into $N =642$ vertices. Each ray corresponding to these vertices was sampled along $c=32$ equidistant locations. Hence, the resolution of the cubic and spherical patches is nearly identical. Both spherical and cubic $f^{\text{in}}_{x, r_i}$ were obtained by trilinearly interpolating the image. To construct $f^{\text{out}}_{x, \text{GT}}$, we determined the ground-truth vessel orientation at a fraction $\eta = 0.25$ of the radius along the centerline, and used $\alpha=3$, $\beta=0.3$ to determine vertex-wise responses (Section \ref{sec:input}).
Furthermore, we created negative samples with a probability of 0.1 to accommodate the data-driven guardrail during tracking (Section \ref{sec:train_strat}).

We divided the three datasets described in Section \ref{sec:data}, into train and test splits before training. For the VMR dataset, we randomly selected 29 cases for training and 11 cases for testing. The ASOCA dataset consisted of 20 healthy and diseased samples, from which we used 16 of each type for training and the remaining four of each type for testing. Lastly, for our AAA dataset, we randomly selected 90 patients for training and used the remaining 18 patients for testing. We refer to these data splits as $\text{VMR}_j, \xspace \text{ASOCA}_j, \xspace \text{AAA}_j, \xspace j\in \{\text{training, test}\}$ respectively.

For all three training sets, we trained each of the architectures mentioned above, resulting in nine different orientation estimators, that we denote by $g_{\{\text{CNN, GAT, GEM}\}, \{\text{VMR, ASOCA, AAA}\}}$. In all three datasets, we rescaled the intensities of all images from the [1200, 200] window/level to intensities within the [0, 1] interval, without clipping, to improve training convergence. During training, we used the scales $R = \{$1, 2, 5, 10, 15, 20, 25, 30, 35, 40, 45, 50$\}$ mm, $R =$ \{1, 2, 5, 7, 10, 15$\}$ mm and $R = \{$5, 10, 15, 20, 25, 30, 35, 40, 45, 50$\}$ mm to train $g_{\{\text{GEM, GAT, CNN}\}, \cdot}$ on VMR, ASOCA and AAA respectively. In addition, we trained the GEM architecture on the three datasets using randomly sampled scales at each training iteration. We denote these SIREs by $\hat{g}_{\text{GEM}, \{\text{VMR, ASOCA, AAA}\}}$, and they were trained on scales randomly sampled from $\mathcal{U}_{[0,50]}, \mathcal{U}_{[0,15]}, \mathcal{U}_{[0,50]}$ respectively.

All experiments were performed on an NVIDIA A40 GPU, with 48GB memory. We trained the orientation estimators for 3,000 epochs. We used an Adam optimiser and learning rates of 0.001 for $\hat{g}_{\text{GEM}, \{ \text{ASOCA, AAA} \}}$, $g_{\text{CNN, VMR}}$, $g_{\text{GAT, ASOCA}}$ and 0.0001 for $\hat{g}_{\text{GEM, VMR}}$, $g_{\text{GEM},\{\text{VMR, ASOCA, AAA}\}}$, $g_{\text{GAT}, \{ \text{VMR, AAA} \}}$ and $g_{\text{CNN}, \{ \text{ASOCA, AAA} \}}$. No data augmentation was used.

\subsection{Centerline tracking}\label{sec:tracking}
We assess the quality of automatically extracted centerlines from AAA\textsubscript{test} and ASOCA\textsubscript{test} using the iterative tracking algorithm. To demonstrate the generalisation of SIRE, we use $g_{\cdot, \text{ASOCA}}$ and $g_{\cdot, \text{AAA}}$ to track AAA and coronary artery centerlines, respectively, despite \textit{not having seen} vessels of this calibre during training. In AAA tracking we also compare the two different GCN architectures, $g_{\text{GEM}, \cdot}$ and $g_{\text{GAT},\cdot}$. Lastly, we assess the effect of using randomly sampled scales during training on automatic extraction of coronary trees.

\subsubsection{AAA tracking}\label{sec:AAA_tracking}
To assess the generalisation of SIRE, we use $g_{ \text{GEM}, \cdot}$ and $g_{\text{GAT}, \cdot}$ trained on the VMR, ASOCA and AAA datasets to track the abdominal aortas of patients in AAA\textsubscript{test}. For training we use the scales $R=\{ 1, 2, 5, 10, 15, 20, 25, 30, 35, 40, 45, 50, 55, 50\}$ mm, $R=\{1, 2, 5, 7, 15 \}$ mm and $R = \{ 5, 10, 15, 20, 25, 30, 35, 40, 45, 50 \}$ mm for VMR, ASOCA and AAA, respectively, (Section \ref{sec:exp}). Next, we manually place a seed point inside the abdominal aorta, at the level of the renal arteries to initiate the tracker. We use scales $R=\{5, 10, 15, ..., 60\}$ mm for $g_{\cdot, \{ \text{VMR, AAA}\}}$ and $R=\{5, 10, 15, ..., 90 \}$ mm for $g_{\cdot, \text{ASOCA}}$ with a step size of $\Delta=0.5$ mm during tracking.

As the ground-truth centerlines of the AAA dataset only cover the abdominal part of the aorta, we use the average inside distance and recall to measure tracking performance. We omit precision and overlap, as SIRE often tracks far beyond the annotated section of the aorta. Figure \ref{fig:AAA_AI} shows recall and average inside distance for the six SIREs. The blue boxplots display the performance of $g_{\text{GEM}, \cdot}$. We observe that recall for tracking AAAs is similar for $g_{\text{GEM}, \cdot}$, regardless of the vessel calibre seen during training. Furthermore, the AI distances are similar for $g_{\text{GEM, ASOCA}}$ and $g_{\text{GEM, AAA}}$ and are slightly larger for $g_{\text{GEM, VMR}}$, likely due to lower centerline quality in VMR\textsubscript{train}. This demonstrates the generalisation of SIRE to vessels of unseen calibre. Moreover, the recall of $g_{\text{GEM}, \cdot}$ is consistently higher than the recall of $g_{\text{GAT}, \cdot}$. However, the AI distances for $g_{\text{GAT}, \cdot}$ are slightly smaller than the AI distance of $g_{\text{GEM}, \cdot}$. In combination with the low recall, this indicates that $g_{\text{GAT},\cdot}$ tends to stop tracking prematurely. In summary, $g_{\text{GEM}, \cdot}$ consistently outperforms $g_{\text{GAT}, \cdot}$ in AAA centerline tracking, and $g_{\text{GEM}, \cdot}$ can generalise to vessels of unseen size while maintaining performance.

Figure \ref{fig:AAA_MPR} shows a multi-planar reconstruction (MPR) of the centerlines obtained from $g_{\text{GEM}, \text{VMR, ASOCA, AAA}}$. This figure also shows the average active scale at each tracking step. Starting at the renal bifurcation, all three trackers continued through the aortic arch until the aortic root. In the opposite direction, tracking continued into the iliac arteries until the scan boundary was reached. The MPRs show the wide range of vessel calibre from the aortic root, through the aneurysm until the common iliac arteries. For all three SIREs the average active scale decreases when the vessel radius decreases. The active scales for $\hat{g}_{\text{GEM,ASOCA}}$ are slightly larger than for $\hat{g}_{\text{GEM}, \{\text{VMR, AAA}\}}$. This suggests that both the selection of the scales, as well as the vessel calibres seen during training influence the linear relation learned between the image features and vessel calibre. Lastly, we observe that $\hat{g}_{\text{GEM}, \{ \text{VMR, ASOCA, AAA} \}}$ continue tracking into the left ventricle. Due to scale-invariance, the locally tubular appearance of the left ventricle and the fact that the left ventricle contains contrast agent, the local entropy remains low. To combat this points from within the left ventricle can be randomly sampled during training, similar to the points outside the lumen. Alternatively, a global context-aware stopping criterion can be considered based on a rough segmentation mask acquired from, e.g., TotalSegmentator \citep{wasserthal_total}.

\begin{figure*}[t!]
    \centering
    \includegraphics[width=\textwidth]{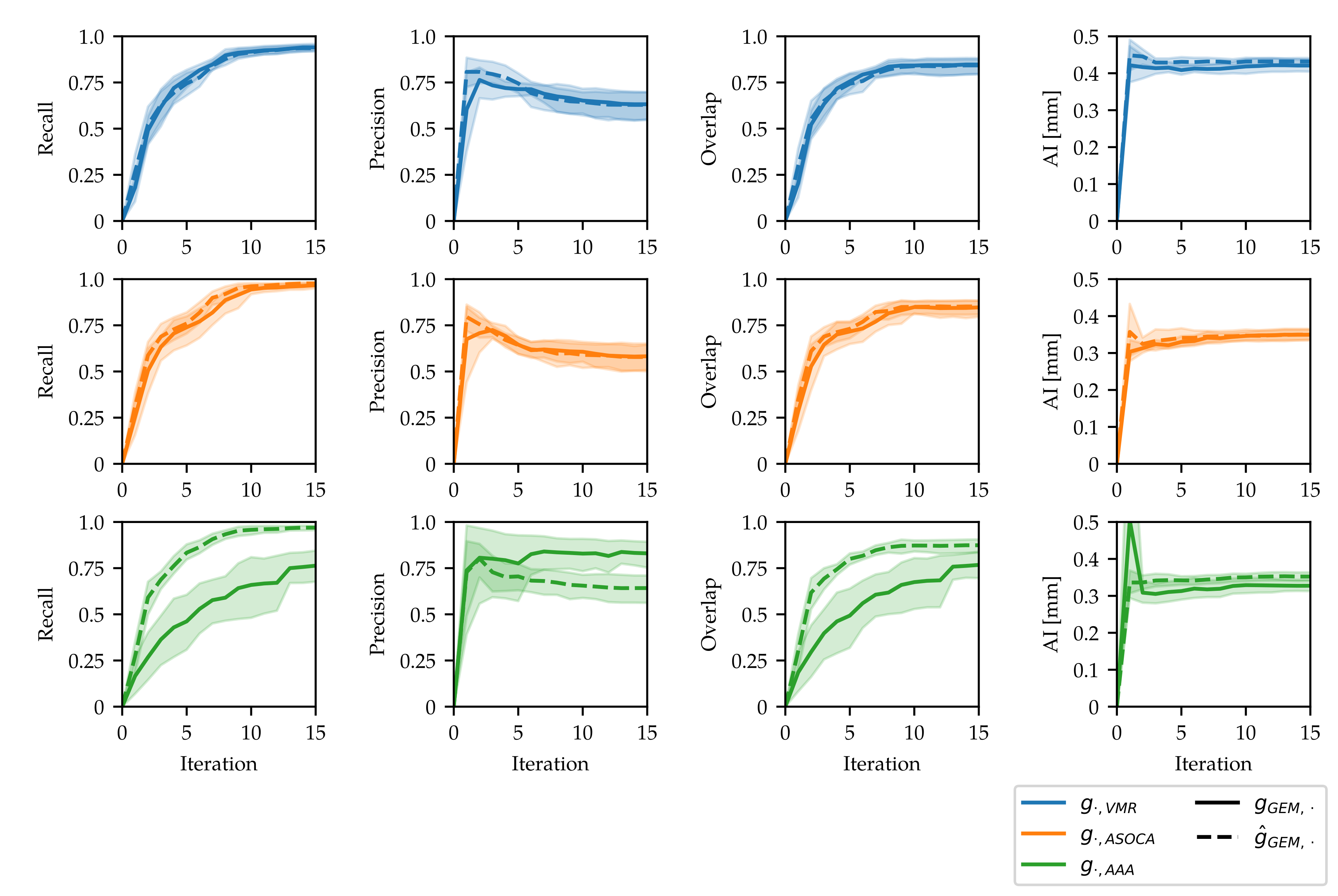}
    \caption{Cumulative recall, precision, overlap and AI distance for the extraction of coronary trees from ASOCA\textsubscript{test} using the six trackers $g_{\text{GEM}, \{\text{VMR, ASOCA, AAA} \} }$, $\hat{g}_{\text{GEM}, \{ \text{VMR, ASOCA, AAA}\}}$. The rows show the performance of SIRE trained on different datasets, all SIREs can extract coronary trees, regardless of their training data.}
    \label{fig:ASOCA_tracking_quantitative}
\end{figure*}

\begin{figure*}[t!]
    \includegraphics[width=\textwidth]{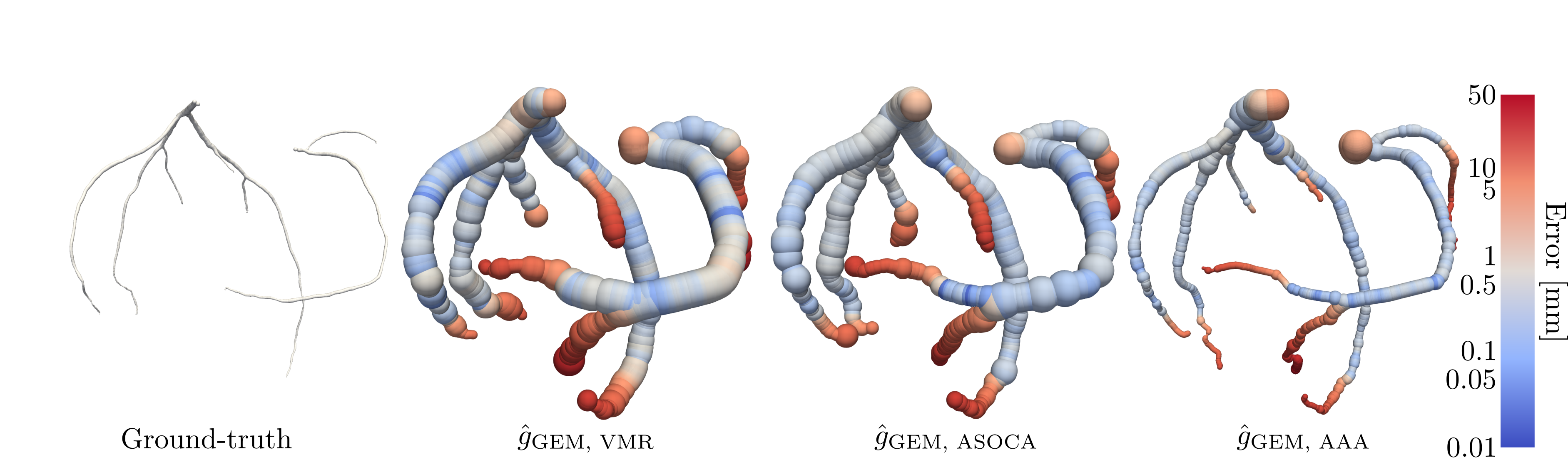}
    \caption{Tracked coronary trees from a patient in ASOCA\textsubscript{test} using $g_{\text{GEM}, \{ \text{VMR, ASOCA, AAA\}}}$ using $R = \{1, 2, 3, 4, 5, 6, 7, 8, 9,  10 \}$, together with the ground-truth vessel tree. Colours represent the distance to the closest ground-truth centerline point, sphere sizes indicate the scale with the highest activation value. Local radii of the ground-truth coronary tree are given as line thickness, on the same scale as the tracked lines. All three trackers continue tracking beyond the ground-truth annotation.}
    \label{fig:ASOCA_track}
\end{figure*}

\subsubsection{Coronary tree extraction}\label{sec:coronary_track}
We evaluated the effect of training on randomly sampled scales on tracking coronary arteries using $g_{\text{GEM}, \{\text{VMR, ASOCA, AAA} \}}$ and $\hat{g}_{\text{GEM}, \{ \text{VMR, ASOCA, AAA}\}}$ trained on the scales described in Section \ref{sec:exp}. To automatically extract the full coronary tree from patients in ASOCA\textsubscript{test}, we first trained an nnU-Net \citep{isensee2021nnu} using ASOCA\textsubscript{train}. Voxelmask segmentations from this nnU-Net had a mean Dice similarity coefficient of 0.86 on ASOCA\textsubscript{test}. Skeletons were acquired from these voxelmasks and used as a queue of seed points as described in Section \ref{sec:vesseltree}. We performed tracking using $R = \{1, 2, 3, 4, 5, 6, 7, 8, 9, 10\}$ mm, a step size of $\Delta=0.25$ mm and a threshold $\tau=0.9$ for the entropy in the stopping criterion for all six SIREs.

After the queue of seed points was empty, we assessed the contribution of each tracked line to the full vessel tree by computing the recall, precision and AI. Figure \ref{fig:ASOCA_tracking_quantitative} shows these cumulative metrics for the trackers trained on the three different datasets and their 95\% confidence intervals. We observe that most trackers can extract coronary trees with a recall of at least 0.9 using at most 15 seeds, regardless of the vessels seen during training. An exception is $g_{\text{GEM, AAA}}$, where recall is low compared to the other five SIREs. The recall curve shows small increments, indicating $g_{\text{GEM, AAA}}$ stops tracking prematurely as its entropy is affected by shifting between AAA\textsubscript{train} and ASOCA\textsubscript{test}. However, training $g_{\text{GEM, AAA}}$ on randomised scales helps with robustness against these shifts, as recall increases close to the level of $g_{\text{GEM, ASOCA}}$.

\begin{figure*}[t!]
    \centering
    \includegraphics[width=\textwidth]{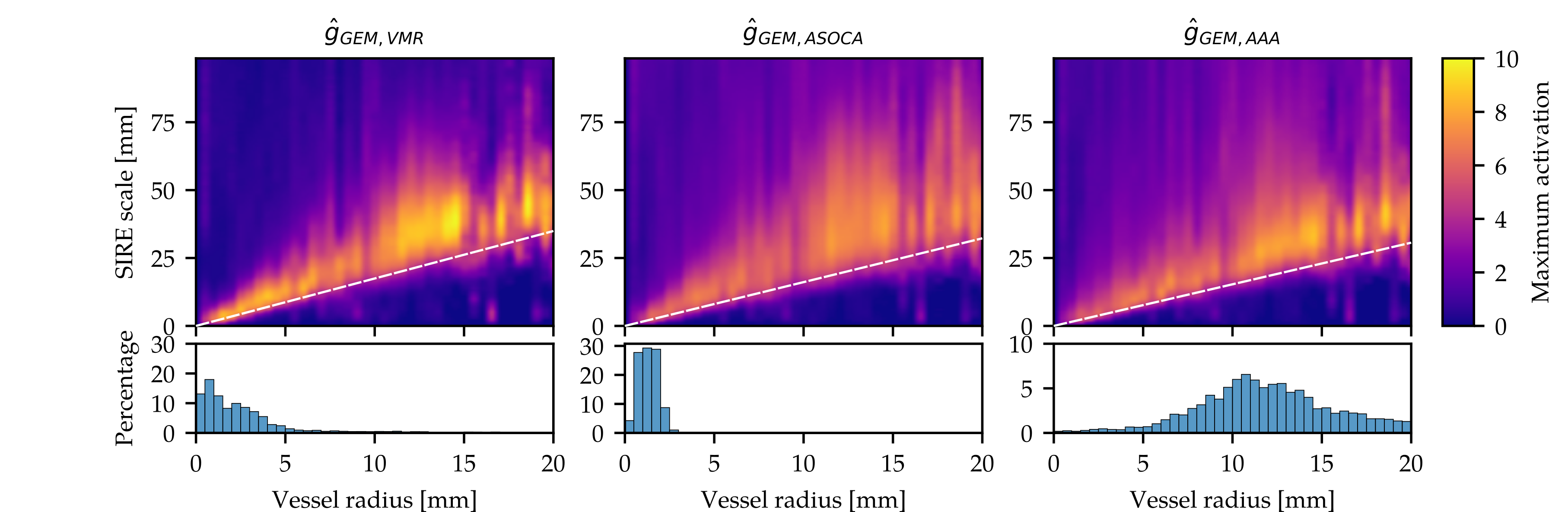}
    \caption{Maximum activations per scale plotted against local vessel radius for $\hat{g}_{\text{GEM}, \{\text{VMR, ASOCA, AAA}\}}$. \textit{Top row}: $\hat{g}_{\text{GEM}}$ trained using scales randomly sampled from $\mathcal{U}_{[0,50]}$, $\mathcal{U}_{[0,15]}$ and $\mathcal{U}_{[0,50]}$ mm for the VMR, ASOCA and AAA dataset, respectively.
    \textit{Bottom row}: Distribution of vessel radii present in the VMR, ASOCA and AAA datasets.}
    \label{fig:scale-invariance}
\end{figure*}

For all six SIREs, a high recall results in a relatively low precision and vice versa. This is because, for the ASOCA dataset, only vessels with a diameter of at least 1 millimetre are annotated. SIRE can track smaller vessels, due to its scale-invariance. Hence, in most cases, tracking often continued far beyond the annotated part of the coronary tree's centerlines, resulting in a high recall, yet low precision. Conversely, low recall and high precision indicate termination of the tracker before the centerline is fully tracked.

The AI distance is between 0.3 and 0.4 millimeters for $g_{\cdot, \{ \text{ASOCA, AAA} \}}$ and between 0.4 and 0.5 millimetres for $g_{\cdot, \text{VMR}}$. In all cases, the AI metrics remain stable throughout tracking, implying that centerline quality is not affected by the different radii and tortuosities of coronary arteries in the tree. Centerlines tracked by $g_{\cdot, \text{VMR}}$ are most likely less accurate due to discrepancies in centerline quality in the training data.

Figure \ref{fig:ASOCA_track} shows the extracted vessel trees for one of the patients in ASOCA\textsubscript{test} using $\hat{g}_{\text{GEM},\{\text{VMR, ASOCA, AAA}\}}$, the distance to the nearest ground-truth centerline point and the most active scale on each of the points passed during tracking. All three SIREs shown here can extract the coronary tree, which is especially remarkable for $g_{\text{GEM, AAA}}$, as it has never seen a coronary artery during training. For all three trackers, we observe that tracked centerlines continue beyond the ground-truth lines, indicated by high errors towards the end of the branches. Moreover, SIRE bases its predictions on spherical input features with a radius larger than the vessel radius, as indicated by the vessel diameters in Figure \ref{fig:ASOCA_track}. $\hat{g}_{\text{GEM}, \{ \text{VMR, ASOCA} \}}$ mostly base their prediction on the same scales, whereas $\hat{g}_{\text{GEM, AAA}}$ bases its predictions on smaller scales.

\subsection{Scale invariance}
We first evaluate the scale-invariant properties of SIRE. During training, SIRE observes vessels with varying diameters. We hypothesize that in a weakly supervised manner the network can learn which scales contain the most relevant information given a vessel of arbitrary size. To test this hypothesis, we used VMR\textsubscript{test}, which has a wide range of vessel radii, as shown in Figure \ref{fig:vis_dataset}. We compared the scale-wise activations $\{f^{\text{out}}_{x,r_i}\}_{r_i\in R}$ of the orientation regressor to the vessel's radius.

We used $\hat{g}_{\text{GEM}, \{\text{VMR, ASOCA, AAA}\}}$ to estimate the vessel orientations, which we trained on scales randomly sampled from $\mathcal{U}_{[0, 50]}$, $\mathcal{U}_{[0,15]}$ and $\mathcal{U}_{[0,50]}$ mm for VMR, ASOCA and AAA respectively (Section \ref{sec:exp}). For this experiment, we grouped centerline points in the VMR dataset based on their local vessel radius, in 0.5 millimeter increments. From each group, we randomly sampled 15 points from which we constructed $\{f^{\text{in}}_{x, r_i}\}_{r_i \in R}$ for $R = \{1, 2, .., 100\}$ and processed these with the three trained networks. The maximum activation at $f^{\text{out}}_{x, r_i}$ for each $r_i$ in $R$ was assessed and averaged for each group of vessel radii.

Figure \ref{fig:scale-invariance} shows the scale-wise activations for SIRE trained on fixed scales together with the distribution of vessel radii in the training data. Although this distribution varies largely between the three datasets, we observe a similar linear trend between the scalewise activations and vessel radius. Moreover, there is an explicit lower bound for when the field of view contains sufficient context to determine the vessel orientation, as indicated by the white dashed line in Figure \ref{fig:scale-invariance}. This bound is consistently larger than the vessel radius, indicating that SIRE requires some context from the vessel's surroundings to determine its orientation. However, the upper bound of high activations is less clearly defined, as the network may still obtain some useful information from relatively large scales. For $\hat{g}_{\text{GEM, VMR}}$ the activation values are higher than for $\hat{g}_{\text{GEM, ASOCA}}$ and $\hat{g}_{\text{GEM, AAA}}$. We suspect that this is related to small discrepancies in contrast levels in the vessels between the datasets, rather than vessel calibre, as this effect is also seen for vessels with radii similar to the training data.

\begin{figure}[t!]
    \centering
    \includegraphics[width=\columnwidth]{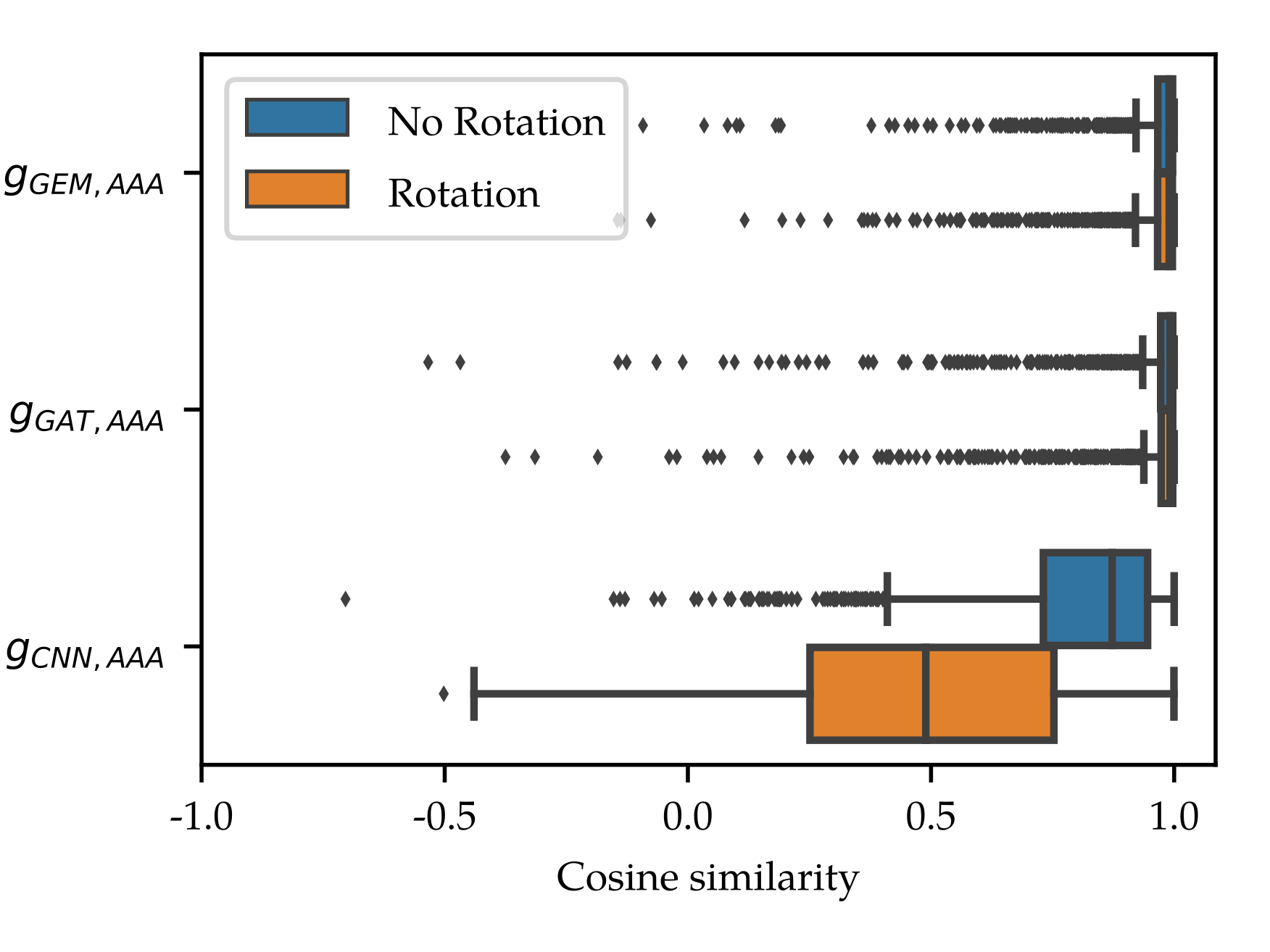}
    \caption{Effect of random rotation of the image data in terms of cosine similarities between the ground-truth vessel directions and the predicted directions for the $g_{\{\text{GEM, GAT, CNN} \},\text{AAA}}$ orientation classifiers on AAA\textsubscript{test}. $g_{\{\text{GEM, GAT}\},\text{AAA}}$ are not affected by random rotations, whereas the performance of $g_{\text{CNN,AAA}}$ drops considerably.}
    \label{fig:plot_rotation_equivariance}
\end{figure}

\begin{figure*}[t!]
    \centering
    \includegraphics[width=\textwidth]{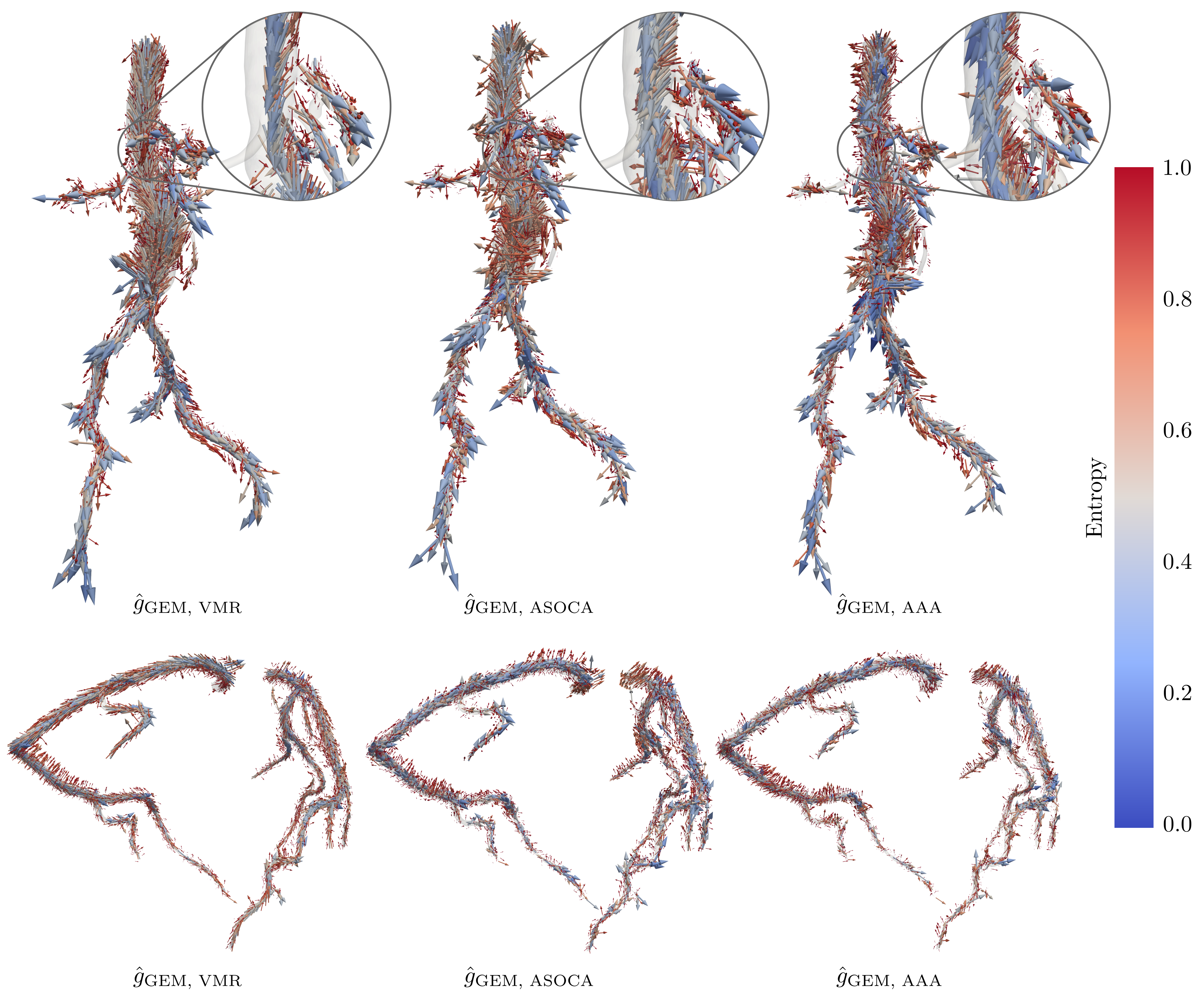}
    \caption{3D visualizations of the predictions and entropies of $\hat{g}_{\text{GEM}, \{\text{VMR, ASOCA, AAA}\}}$, on points sampled in and around the vessel segmentations of a patient from VMR\textsubscript{test} \textit{(top row)} and ASOCA\textsubscript{test} \textit{(bottom row)}. Colours indicate the entropy, arrow directions the predicted direction closest to the positive vertical direction, and arrow size represents the maximum activation on $f^{\text{out}}_{x, \text{max}}$.}
    \label{fig:entropy_maps}
\end{figure*}

\subsection{Rotation equivariance}\label{sec:rot_eq}
We assess the accuracy of the vessel orientation estimation and SO(3)-equivariance of SIRE, using $g_{\{\text{GEM,GAT,CNN}\},\text{AAA}}$, i.e. the GEM, GAT and CNN architectures trained on the AAA dataset using fixed scales. For each patient in AAA\textsubscript{test}, we randomly sample 50 points on the centerline and extract $\{f^{\text{in}}_{x, r_i}\}_{r_i \in R}$ at scales $R = \{ 5, 10, 15, 20, 25, 30, 35, 40, 45, 50\}$. These samples are processed twice by each trained model: with and without random rotations in $\mathbb{R}^3$. We infer the predicted directions from the network outputs and calculate the cosine similarity to the ground-truth directions. The vessel orientation in the AAA dataset is homogeneous in the craniocaudal direction (Fig. \ref{fig:vis_dataset}). Hence, any model that is SO(3)-equivariant, will be unaffected by random rotation of samples at inference, although these orientations were unseen during training.

Figure \ref{fig:plot_rotation_equivariance} shows the results of this experiment. The blue boxplots indicate the cosine similarity for samples not rotated at inference. We observe that $g_{\text{GEM, AAA}}$ and $g_{\text{GAT, AAA}}$, the two models based on graph convolution on a sphere, both have a median cosine similarity of 0.99. We observe some outliers, that are mostly explained by SIRE having higher activations around bifurcating vessels in some cases. $g_{\text{CNN, AAA}}$, the model based on a 3D CNN as in \cite{wolterink2019coronary,gao2021joint,salahuddin2021multi,van2022untangling,su2023deep}, has a median cosine similarity of 0.87. This indicates that $g_{\text{CNN}}$ performs worse at estimating the local vessel orientation than $g_{\{ \text{GEM, GAT}\}}$, even when the sample orientation is similar to those seen during training. The orange boxplots show the cosine similarities of the predicted directions for samples randomly rotated at inference. We verify that, indeed, the performance of $g_{\{\text{GEM, GAT}\},\text{AAA}}$, is unaffected by these random rotations. The cosine similarity of $g_{\text{CNN, AAA}}$ drops rapidly to a median value of 0.49 when random rotations are applied during inference, implying that this model depends on orientation information and is thus not SO(3)-equivariant. In summary, processing the spherical image volumes projected to the spherical surface, as we propose in this work and is implemented in $g_{\{\text{GEM, GAT}\},\text{AAA}}$ is indeed SO(3)-equivariant, conform definition \ref{def:equivariance} and the commutative diagram in Figure \ref{fig:commutative_diagram}.

\begin{figure}[t!]
    \centering
    \includegraphics[width=\columnwidth]{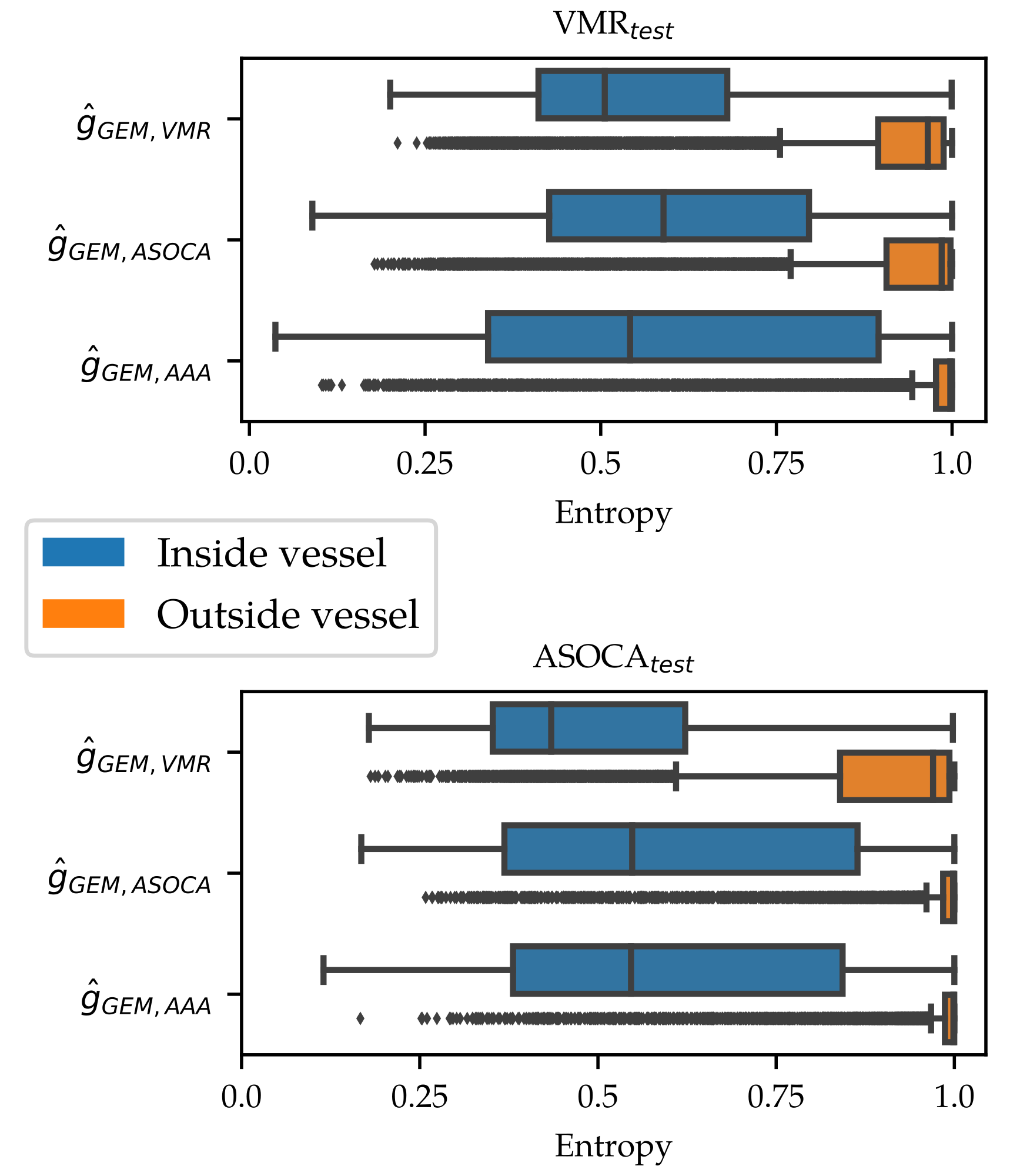}
    \caption{Boxplots showing entropy values for predictions from  and $\hat{g}_{\text{GEM}, \{ \text{VMR, ASOCA, AAA} \}}$ at locations in- and outside the vessel for the patients from VMR and ASOCA shown in Figure \ref{fig:entropy_maps}.}
    \label{fig:entropy_boxplot}
\end{figure}

\subsection{Model uncertainty}\label{sec:entropy}
The entropy of $\sigma\left(f^{\text{out}}_{x, \text{max}}\right)$ is used as a surrogate for the model's uncertainty and stopping criterion in the iterative tracking algorithm, where $\sigma$ represents the softmax function used to transform $f^{\text{out}}_{x, \text{max}}$ into a probability distribution. To verify that entropy is indeed low inside the vessel lumen and high outside the vessel, we sample points for patients in VMR\textsubscript{test} and ASOCA\textsubscript{test} using their ground-truth vessel segmentation masks. At each point, we constructed multiscale samples at scales $R = \{ 1, 2, 5, 10, 15, 20, 25, 30, 35, 40, 45, 50, 55, 60 \}$ mm and $R = \{1, 2, 5, 7, 10\}$ mm, for VMR\textsubscript{test} and ASOCA\textsubscript{test}, respectively. Subsequently, we processed these using  $\hat{g}_{\text{GEM}, \{ \text{VMR, ASOCA, AAA} \}}$ and determined entropy, predicted directions and the maximum activation values on $f^{\text{out}}_{x, \text{max}}$.

Figure \ref{fig:entropy_maps} contains an example of the entropy values, where the arrow directions, colours and sizes indicate one of the predicted directions, entropy value and maximum activation values respectively. This Figure shows that the entropy is generally low in the vessel lumen, and increases towards the vessel boundary, for the networks trained on all three datasets. Note that entropy in the iliac and renal arteries is low for $\hat{g}_{\text{GEM, ASOCA}}$, while this model has only seen coronary arteries during training. Conversely, for $\hat{g}_{\text{GEM, AAA}}$, entropy is also low inside the coronary artery lumen for the patient in ASOCA\textsubscript{test}, while this network has only seen AAAs during training. We also observe low entropies inside the iliac arteries. In summary, entropy is a suitable stopping criterion, as its value differs inside and outside the vessel lumen.

A quantitative overview of these entropy values for $\hat{g}_{\text{GEM}, \{ \text{VMR, ASOCA, AAA} \}}$ on VMR\textsubscript{test} and ASOCA\textsubscript{test} is given in Figure \ref{fig:entropy_boxplot}. For all three SIREs evaluated on these two datasets, the median entropy values are indeed lower inside the vessel than outside the vessel, regardless of the vessels seen during training. However, the standard deviation of entropy inside the vessel is large, hence sometimes the entropy inside the vessel is close to the entropy outside the vessel. This may cause early termination of the tracker when using entropy as a stopping criterion.

\section{Discussion}
We have introduced SIRE: a scale-invariant, rotation-equivariant method to estimate local vessel orientations based on 3D image data. SIRE consists of a graph neural network, operating on the spherical surface on which multi-scale spherical image volumes have been projected. We have shown that SIRE adapts to vessels of sizes unseen during training. Tracking with SIRE trained on coronaries obtains similar results when tracking coronaries compared to a tracker with SIRE trained on AAAs, and vice versa; tracking AAAs using SIRE trained on coronary arteries achieves similar results in similar results when using SIRE trained on AAAs. SIRE has the potential to facilitate downstream tasks in the management of patients with cardiovascular diseases, such as measuring vessel diameters, plaque thickness, or calcifications. 

SIRE is highly modular, and in our experiments, we compared the use of three different network architectures at its core: CNN, GAT \citep{brody2021attentive}, and GEM-CNN\citep{dehaan2021}. The GAT and GEM-CNN networks are GCNs and use spherical multi-scale input patches as input. In contrast, the CNN operated on canonically oriented cubic patches. We found that the CNN was unable to simultaneously learn to distinguish scale importance and vessel orientations, which was reflected in poor estimations of the local vessel orientations. Among the GCNs, we found that the GEM-CNN outperformed the GAT network in AAA tracking. This performance gap can be explained by the expressiveness of the networks: GEM convolutions distinguish neighbours based on their relative positions, whereas the convolutions in GAT do not have this property. For the image processing task handled in SIRE, kernel expressiveness turns out to be essential. This confirms the findings of previous work in which a direct comparison has been made between GEM-CNN and GAT~\citep{suk2022mesh}.

SIRE exploits rotation and scale symmetries in data. We show that this allows us to generalise well between vessels with different scales and levels of tortuosity visualized in CT, but there are two caveats. First, because we operate on discrete 3D volumes and triangular meshes, the symmetry-preserving properties of SIRE are limited to a discrete set of rotations and scale transformations~\citep{edixhoven2023using}. This effect can be partly overcome by using a higher resolution in scales sampled during training, as well as a higher vertex resolution in the icosphere $\mathcal{M}$. Second, SIRE is not domain invariant, i.e., it does not exploit symmetry groups that affect texture, and will thus not generalise to MRI or US images. In future work, this problem could be mitigated by domain adaptation approaches~\citep{guan2021domain}, ideally without the need for additional centerline annotations in the target domain. However, texture invariance might not always be desirable and might make it challenging to distinguish between, e.g., arteries and veins.

We have here integrated SIRE into a relatively simple tracking algorithm, with good results. The tracker follows the most prominent direction, and uses a stopping criterion to terminate. This stopping criterion was here based on the entropy on $f^{\text{out}}_{x, \text{max}}$. The main advantages of using entropy are its boundedness and ease of use, but other uncertainty estimation approaches could be used in SIRE, such as Monte-Carlo dropout \citep{gal2016dropout} or the eigenvalue analysis of the local structure tensor at the most active scale \citep{kumar20133d}. SIRE is a general orientation estimator and can be integrated into any tracking algorithm. For example, the hyperparameters of the tracker, which were now mostly found empirically, could be set using reinforcement learning \citep{su2023deep}. Moreover, while we optimized the model locally based on individual points along the centerline, SIRE could be integrated into a recursive tracker that provides centerlines, allowing training with centerline-based loss function, such as clDice \citep{shit2021cldice}. Additionally, in combination with multiple hypothesis tracking, SIRE could be adapted to track full vessel trees \citep{friman2010multiple}.

SIRE is very data-efficient, and through the use of symmetries only requires a relatively modest \textit{amount} of training data. In addition, we found that using randomly sampled scales during training is beneficial for generalisation. However, we also noticed that the \textit{quality} of training data can substantially affect the performance of SIRE. The reference centerline annotations used in ASOCA and AAA were of higher quality than those in VMR - where accurate centerlines are not a goal - and this was reflected in the performance of the method when trained with these data sets. This highlights the importance of high-quality public datasets such as \cite{gharleghi2023annotated}. 

\section{Conclusion}

In conclusion, we presented a modular scale-invariant, rotation equivariant method that determines the local vessel orientation for vessels of any size or tortuosity. SIRE is flexible and can be used as a separate module in any vessel tracking algorithm. Because of its generalisation, SIRE has the potential to automate centerline extraction for many downstream analysis tasks in vessels of any calibre and tortuosity, impacting clinical practice, cardiovascular modelling and image processing.

\section*{Acknowledgments}
Jelmer M. Wolterink was supported by the NWO domain Applied and Engineering Sciences VENI grant (18192). This work is part of the 4TU Precision Medicine
program supported by High Tech for a Sustainable Future, a framework commissioned by the four Universities of Technology of the Netherlands.

Part of the data used in this work was provided in whole or in part with
Federal funds from the National Library of Medicine under Grant
 No. R01LM013120, and the National Heart, Lung, and Blood Institute,
 National Institutes of Health, Department of Health and Human Services, 
under Contract No. HHSN268201100035C. % statement VMR

\section*{Ethical approval}
Our in-house dataset containing CTA scans of patients with abdominal aortic aneurysms was in accordance with the ethical standards of the institutional and/or national research committee and with the 1964 Helsinki declaration and its later amendments or comparable ethical standards. The study was approved by the local ethical committee of Amsterdam UMC (VUMC2020.323). For this type of study formal consent was not required.

% \section*{References}

% Please ensure that every reference cited in the text is also present in
% the reference list (and vice versa).https://www.overleaf.com/project/63fdffecf8cffd2092802b32

% \section*{\itshape Reference style}

% Text: All citations in the text should refer to:
% \begin{enumerate}
% \item Single author: the author's name (without initials, unless there
% is ambiguity) and the year of publication;
% \item Two authors: both authors' names and the year of publication;
% \item Three or more authors: first author's name followed by `et al.'
% and the year of publication.
% \end{enumerate}
% Citations may be made directly (or parenthetically). Groups of
% references should be listed first alphabetically, then chronologically.

\bibliographystyle{model2-names.bst}\biboptions{authoryear}
\bibliography{refs}

\end{document}